\documentclass[10pt,journal,compsoc]{IEEEtran}
\ifCLASSOPTIONcompsoc
  \usepackage[nocompress]{cite}
\else
  \usepackage{cite}
\fi
\ifCLASSINFOpdf
\else
\fi
\usepackage[utf8]{inputenc} 
\usepackage[T1]{fontenc}    
\usepackage{hyperref}       
\usepackage{url}            
\usepackage{booktabs}       
\usepackage{amsmath}		
\usepackage{amssymb}		
\usepackage{MnSymbol}
\usepackage{amsfonts}       
\usepackage{gensymb}		
\usepackage{nicefrac}       
\usepackage{microtype}      
\usepackage{graphicx}		
\usepackage{dblfloatfix}
\usepackage{float}
\usepackage{color}			
\usepackage[colorinlistoftodos]{todonotes}
\usepackage{caption}
\usepackage{subcaption}
\usepackage{multirow}
\usepackage{makecell}
\usepackage{lipsum}
\usepackage{comment}

\usepackage{diagbox, eqparbox}

\makeatletter
\renewcommand{\paragraph}{%
  \@startsection{paragraph}{4}%
  {\z@}{2ex \@plus 1ex \@minus .2ex}{-1em}%
  {\normalfont\normalsize\bfseries}%
}
\makeatother

\makeatletter
\newcommand{\xdashrightarrow}[2][]{\ext@arrow 0359\rightarrowfill@@{#1}{#2}}
\def\rightarrowfill@@{\arrowfill@@\relax\relbar\rightarrow}
\def\leftarrowfill@@{\arrowfill@@\leftarrow\relbar\relax}
\def\leftrightarrowfill@@{\arrowfill@@\leftarrow\relbar\rightarrow}
\def\arrowfill@@#1#2#3#4{%
  $\m@th\thickmuskip0mu\medmuskip\thickmuskip\thinmuskip\thickmuskip
   \relax#4#1
   \xleaders\hbox{$#4#2$}\hfill
   #3$%
}

\setcellgapes{3pt}

\newcolumntype{?}[1]{!{\vrule width #1}}
\newcommand{\hbline}{\Xhline{2.5\arrayrulewidth}}

\newcommand{\defaulttabcolsep}{\tabcolsep}

\newlength{\Oldarrayrulewidth}

\makeatletter
\newcommand{\citem}[1]{%
  \begingroup\let\cite@adjust\@empty
  \cite{#1}%
  \endgroup
}

\begin{document}
%
\title{Unified Adversarial Invariance}
%
%
%
%

        
\author{Ayush~Jaiswal,~\IEEEmembership{Student~Member,~IEEE,}
        Yue~Wu,~\IEEEmembership{Member,~IEEE,}
        Wael~AbdAlmageed, 
        and~Premkumar~Natarajan,~\IEEEmembership{Senior~Member,~IEEE}
%
%
\thanks{A. Jaiswal, Y. Wu, W. AbdAlmageed, and P. Natarajan are with the Information Sciences Institute, University of Southern California, Marina del Rey, USA.\protect\\
E-mail: \{ajaiswal, yue\_wu, wamageed, pnataraj\}@isi.edu\protect\\
\protect\\
Extension of our previous work ``Unsupervised Adversarial Invariance'' published in Advances in Neural Information Processing Systems, 2018\protect\\
\protect\\
\copyright 2019 IEEE. Personal use of this material is permitted. Permission from IEEE must be obtained for all other uses, in any current or future media, including reprinting/republishing this material for advertising or promotional purposes, creating new collective works, for resale or redistribution to servers or lists, or reuse of any copyrighted component of this work in other works.
}}

\IEEEtitleabstractindextext{%
\begin{abstract}
We present a unified invariance framework for supervised neural networks that can induce independence to nuisance factors of data without using any nuisance annotations, but can additionally use labeled information about biasing factors to force their removal from the latent embedding for making fair predictions. Invariance to nuisance is achieved by learning a split representation of data through competitive training between the prediction task and a reconstruction task coupled with disentanglement, whereas that to biasing factors is brought about by penalizing the network if the latent embedding contains any information about them. We describe an adversarial instantiation of this framework and provide analysis of its working. Our model outperforms previous works at inducing invariance to nuisance factors without using any labeled information about such variables, and achieves state-of-the-art performance at learning independence to biasing factors in fairness settings.
\end{abstract}

\begin{IEEEkeywords}
Invariance, Fairness, Disentanglement, Representation Learning, Adversarial Learning, Deep Neural Networks.
\end{IEEEkeywords}}

\maketitle

\IEEEdisplaynontitleabstractindextext

%
\IEEEpeerreviewmaketitle

\IEEEraisesectionheading{\section{Introduction}\label{sec:introduction}}

\IEEEPARstart{A}{} common formulation of supervised machine learning is the estimation of the conditional probability $p(y|x)$ from data where $x$ and $y$ denote data samples and target variables, respectively. This involves the decomposition of $x$ into its underlying factors of variation, such that associations can be learned between $y$ and the said factors to approximate a mapping from $x$ to $y$. However, trained models often learn to incorrectly associate $y$ with nuisance factors of data, which are truly irrelevant to the prediction of $y$, leading to overfitting and poor generalization on test cases that contain unseen variations of such factors. For example, a nuisance variable in the case of face recognition is the lighting condition in which the photograph was captured. A recognition model that associates lighting with subject identity is expected to perform poorly.



Developing machine learning methods that are invariant to nuisance factors has been a long-standing problem; studied under various names such as feature selection~\cite{bib:feat_select_survey}, robustness through data augmentation~(\citem{bib:data_aug_1,bib:data_aug_2,bib:data_aug_3}) and invariance induction~(\citem{bib:uai,bib:infodropout,bib:vib}). An architectural solution to this problem for deep neural networks (DNN) is creation of neural network units that capture specific forms of information, and thus are inherently invariant to certain nuisance factors~\cite{bib:drl}. For example, convolutional operations coupled with pooling strategies capture shift-invariant spatial information while recurrent operations robustly capture high-level trends in sequential data. However, this approach requires significant effort for \emph{engineering} custom modules and layers to achieve invariance to \emph{specific} nuisance factors, making it inflexible. A different but popularly adopted solution to the problem of nuisance factors is the use of data augmentation where synthetic versions of real data samples are generated, during training, with \emph{specific} forms of variation~(\citem{bib:drl,bib:face_rec}). For example, rotation and translation are typical methods of augmentation used in computer vision, especially for classification and detection tasks. However, models trained na\"ively on the augmented dataset become robust to limited forms of nuisance by learning to associate every seen variation of such factors to the target. Consequently, such models perform poorly when applied to data exhibiting unseen variations of those nuisance variables, e.g., images of objects at previously unseen orientations or colors in the case of object detection. Thus, na\"ively training with data augmentation makes models \emph{partially} invariant to the variables \emph{accounted for in the augmentation process}.

Furthermore, training datasets often contain factors of variation that are correlated with the prediction target but should not be incorporated in the prediction process to avoid skewed decisions that are unfair to under-represented categories of these \textit{biasing} factors. This can also be viewed as a ``class-imbalance'' problem with respect to the biasing factor instead of the target variable. For example, gender and race are biasing factors in many human-centric tasks like face recognition~\cite{bib:bias_face_recognition}, sentiment analysis~\cite{bib:bias_sentiment_analysis}, socio-economic assessments~\cite{bib:bias_socioeconomic}, etc. Models that do not account for such bias make incorrect predictions and can sometimes be unethical to use. It is, therefore, necessary to develop mechanisms that train models to be invariant to not only nuisance but also biasing factors of data.

Within the framework of DNNs, predictions can be made invariant to undesired (nuisance or biasing) factors $z$ if the latent representation of data learned by a DNN at any given layer does not contain any information about those factors. This view has been adopted by recent works as the task of invariant representation learning (\citem{bib:infodropout,bib:vib,bib:lfr,bib:nnmmd,bib:vfae,bib:sai,bib:cvib,bib:hcv}) through specialized training mechanisms that encourage the exclusion of undesired variables from the latent embedding. Models trained in this fashion to be invariant to nuisance variables $z$, as opposed to training simply with data augmentation, become robust by exclusion rather than inclusion. Therefore, such models are expected to perform well even on data containing variations of specific nuisance factors that were not seen during training. For example, a face recognition model that learns to not associate lighting conditions with the identity of a person is expected to be more robust to lighting conditions than a similar model trained na\"ively on images of subjects under \emph{certain} different lighting conditions~\cite{bib:sai}. Similarly, the use of such mechanisms to train models to be invariant to biasing $z$ provides better guarantees that the sensitive information is not incorporated in the prediction process (\citem{bib:nnmmd,bib:vfae}).

Invariant representation learning methods can be broadly categorized into two classes --- (1) those that do not employ annotations of undesired $z$ for learning invariance and (2) those that do. The former class of methods is better suited for nuisance $z$ ($z \perp y$) than the latter intuitively because it does not require labeled-information or domain knowledge of the possible nuisance factors and their variations and is, in theory, capable of learning invariance to all nuisance $z$ jointly~\cite{bib:achille2018}. However, these methods cannot be used in fairness settings, where the biasing $z$ is correlated with the target $y$, because they can only discard $z$ that are \emph{not} correlated with $y$ (i.e., nuisance). Hence, it is necessary to employ the second class of methods in fairness settings.

We present a unified framework for invariance induction that can be used without $z$-labels for robustness to nuisance and \emph{additionally} with $z$-annotations for independence to biasing factors. The framework promotes invariance to nuisance through separating the underlying factors of $x$ into two latent embeddings --- $e_1$, which contains all the information required for predicting $y$, and $e_2$, which contains information irrelevant to the prediction task. While $e_1$ is used for predicting $y$, a noisy version of $e_1$, denoted as $\tilde{e}_1$, and $e_2$ are used to reconstruct $x$. This creates a \textit{competitive} scenario where the reconstruction module tries to pull information into $e_2$ (because $\tilde{e}_1$ is unreliable) while the prediction module tries to pull information into $e_1$. The training objective is augmented with a disentanglement loss that penalizes the model for overlapping information between $e_1$ and $e_2$, futher boosting the competition between the prediction and reconstruction tasks. In order to deal with known biasing factors $z$ of data, a proxy loss term for the mutual information $I(e_1:z)$ is added to the training objective, creating a framework that learns invariance to both nuisance and biasing factors. We present an adversarial instantiation of this generalized formulation of the framework, where disentanglement is achieved between $e_1$ and $e_2$ in a novel way through two adversarial \textit{disentanglers} --- one that aims to predict $e_2$ from $e_1$ and another that does the inverse, and invariance to biasing $z$ is achieved through an adversarial $z$-discriminator that aims to predict $z$ from $e_1$. The parameters of the combined model are learned through adversarial training between (a) the encoder, the predictor and the decoder, and (b) the disentanglers (for both nuisance and biasing factors) and the $z$-discriminator (for biasing factors).

The framework makes no assumptions about the data, so it can be applied to any prediction task without loss of generality, be it binary/multi-class classification or regression. We provide results on five tasks involving a diverse collection of datasets -- (1) invariance to inherent nuisance factors, (2) effective use of synthetic data augmentation for learning invariance, (3) learning invariance to arbitrary nuisance factors by leveraging Generative Adversarial Networks (GANs)~\cite{bib:gan}, (4) domain adaptation, and (5) invariance to biasing factors for fair representation learning. Our framework outperforms existing approaches on all of these tasks. This is especially notable for invariance to nuisance in tasks (1) and (2) where previous state-of-the-art works incorporate $z$-labels whereas our model is trained without these annotations.


The rest of the paper is organized as follows. Section~\ref{sec:related_work} discusses related work on invariant representation learning. In Section~\ref{sec:uaif} we describe our unified adversarial invariance framework. Analysis of the model is provided in Section~\ref{sec:theoretical} and results of empirical evaluation in Section~\ref{sec:evaluation}. Finally, Section~\ref{sec:conclusion} concludes the paper.

\section{Related Work}
\label{sec:related_work}

Methods for preventing supervised models from learning false associations between target variables and nuisance factors have been studied from various perspectives including feature selection~\cite{bib:feat_select}, robustness through data augmentation (\citem{bib:data_aug_1,bib:data_aug_2,bib:data_aug_3}) and invariance induction (\citem{bib:infodropout,bib:vib,bib:lfr,bib:nnmmd,bib:vfae,bib:sai,bib:cvib,bib:hcv}). Feature selection has typically been employed when data is available as a set of conceptual features, some of which are irrelevant to the prediction tasks. Our approach learns a split representation of data as $e = [e_1 \ e_2]$ where $e_1$ contains factors that are relevant for $y$-prediction and $e_2$ contains nuisance variables. This can be interpreted as an implicit feature selection mechanism for neural networks, which can work on both raw data (such as images) and feature-sets (e.g., frequency features computed from raw text). Popular feature selection methods~\cite{bib:feat_select} incorporate information-theoretic measures or use supervised methods to score features with their importance for the prediction task and prune the low-scoring features. Our framework performs this task implicitly on latent features that the model learns by itself from the provided data.

Deep neural networks (DNNs) have outperformed traditional methods at several supervised learning tasks. However, they have a large number of parameters that need to be estimated from data, which makes them especially vulnerable to learning relationships between target variables and nuisance factors and, thus, overfitting. The most popular approach to solve this has been to expand the data size and prevent overfitting through synthetic data augmentation, where multiple copies of data samples are created by altering variations of certain known nuisance factors. DNNs trained with data augmentation have been shown to generalize better and be more robust compared to those trained without augmentation in many domains including vision~(\citem{bib:face_rec,bib:sbmrinet,bib:data_aug_2}), speech~\cite{bib:data_aug_1} and natural language~\cite{bib:data_aug_3}. This approach works on the principle of inclusion, in which the model learns to associate multiple seen variations of those nuisance factors to each target value. In contrast, our method encourages exclusion of information about nuisance factors from latent features used for predicting the target, thus creating more robust representations. Furthermore, combining our method with data augmentation additionally helps our framework remove information about nuisance factors used to synthesize data, without the need to explicitly quantify or annotate the generated variations. This is especially helpful in cases where augmentation is performed using sophisticated analytical or composite techniques~\cite{bib:face_rec}.

Information bottleneck~\cite{bib:ib} has been widely used to model unsupervised methods of invariance to nuisance variables within supervised DNNs in recent works (\citem{bib:vib,bib:cvib,bib:achille2018}). The working mechanism of these methods is to minimize the mutual information of the latent embedding $h$ and the data $x$, i.e., $I(x:h)$, while maximizing $I(h:y)$ to ensure that $h$ is maximally predictive of $y$ but a minimal representation of $x$ in that regard. Hence, these methods compress data into a compact representation and indirectly minimize $I(h:z)$ for nuisance $z \perp y$. An optimal compression of this form would get rid of all such nuisance factors with respect to the prediction target~\cite{bib:achille2018}. However, the bottleneck objective is difficult to optimize~(\citem{bib:infodropout,bib:vib}) and has consequently been approximated using variational inference in prior work~\cite{bib:vib}. Information Dropout~\cite{bib:infodropout}, which is a data-dependent generalization of dropout~\cite{bib:dropout}, also optimizes the bottleneck objective indirectly. In contrast to these methods of learning nuisance-free representations through explicit compression, our framework learns a split representation of data into an \textit{informative} embedding that is relevant for $y$-prediction and a \textit{nuisance} embedding by encouraging the separation of these factors of data within neural networks. The competing objectives of prediction and reconstruction coupled with the orthogonality constraint in our framework indirectly, yet intuitively, optimize the bottleneck objective with respect to the hidden representation $e_1$ and the prediction target $y$ by requiring $e_1$ to hold only those factors that are \emph{essential} for predicting $y$ while pushing all other factors of data (nuisance) into $e_2$ such that the decoder has more direct access to such information and can better reconstruct $x$.

Several supervised methods for invariance induction have also been developed recently (\citem{bib:sai,bib:lfr,bib:vfae,bib:nnmmd,bib:cvib}). These methods use annotations of unwanted factors of data within specialized training mechanisms that force the removal of these variables from the latent representation. Zemel et al.~\cite{bib:lfr} learn fair representations by optimizing an objective that maximizes the performance of $y$-prediction while enforcing group fairness through statistical parity. Maximum Mean Discrepancy (MMD)~\cite{bib:mmd} has been used directly as a regularizer for neural networks in the NN+MMD model of~\cite{bib:nnmmd}. The Variational Fair Autoencoder (VFAE)~\cite{bib:vfae} optimizes the information bottleneck objective indirectly in the form of a Variational Autoencoder (VAE)~\cite{bib:vae} and uses MMD to to boost the removal of unwanted factors from the latent representatoin. The Hilbert-Schmidt Information Criterion (HSIC)~\cite{bib:hsic} has been used similarly in the HSIC-constrained VAE (HCV)~\cite{bib:hcv} to enforce independence between the intermediate hidden embedding and the undesired variables. Moyer et al.~\cite{bib:cvib} achieve invariance to $z$ by augmenting the information bottleneck objective with the mutual information between the latent representation and $z$, and optimizing its variational bound (Conditional Variational Information Bottleneck or CVIB). Such methods are expected to more explicitly remove \emph{certain specific} nuisance factors of data from the latent representation as compared to the aforementioned unsupervised methods. A shortcoming of this approach is the requirement of domain knowledge of possible nuisance factors and their variations, which is often hard to find~\cite{bib:drl}. Additionally, this solution applies only to cases where annotated data is available for each nuisance factor, such as labeled information about the lighting condition of each image in the face recognition example, which is often not the case. However, supervised methods are well-suited for inducing invariance to biasing factors of data, which are correlated with the prediction target $y$ but are unfair to under-represented groups within the training set, e.g., age, gender, race, etc. in historical income data. This is because the correlation of biasing factors with the prediction target makes it impossible for unsupervised invariance methods to automatically remove them from the latent representation, and external information about these variables is, hence, required. Our framework can use annotations of biasing factors to learn invariance to them. Hence, it is suited for fair representation learning, as well.

Disentangled representation learning is closely related to our work since disentanglement is one of the pillars of invariance induction in our framework as the model learns two embeddings (for any given data sample) that are expected to be uncorrelated to each other. Our method shares some properties with multi-task learning (MTL)~\cite{bib:mtl} in the sense that the model is trained with multiple objectives. However, a fundamental difference between our framework and MTL is that the latter promotes a shared representation across tasks whereas the only information shared \emph{loosely} between the tasks of predicting $y$ and reconstructing $x$ in our framework is a noisy version of $e_1$ to help reconstruct $x$ when combined with a separate encoding $e_2$, where $e_1$ itself is used directly to predict $y$.


\section{Unified Adversarial Invariance}
\label{sec:uaif}



We present a generalized framework for induction of invariance to undesired (both nuisance and biasing) factors $z$ of data, where $z$ information is not necessary for the exclusion of nuisance but is employed for making $y$-predictions independent of biasing factors. The framework brings about invariance to nuisance $z$ by disentangling information required for predicting $y$ from other unrelated information contained in $x$ through the incorporation of data reconstruction as a competing task for the primary prediction task. This is achieved by learning a split representation of data as $e = [e_1 \ e_2]$, such that information essential for the prediction task is pulled into $e_1$ while all other information about $x$ migrates to $e_2$. In order to further learn invariance to \emph{known} biasing $z$, the training objective of the framework penalizes the model if the encoding $e_1$ contains any information about these $z$. We present an adversarial instantiation of this framework --- Unified Adversarial Invariance (UnifAI), which treats disentanglement of $e_1$ and $e_2$, and removal of biasing $z$ from $e_1$ as adversarial objectives with respect to the competitive prediction and reconstruction tasks.

\setlength{\tabcolsep}{0.5em} 
\begin{table}
\makegapedcells
\centering
\caption{\label{tab:notation}Key Concepts and Framework Components}
\begin{tabular}{ l ?{1.5pt} l }
  \hbline
  \multicolumn{1}{c?{1.5pt}}{\textbf{Term}} & \multicolumn{1}{c}{\textbf{Meaning}} \\
  \hbline
  $x$ & Data sample \\
  $y$ & Prediction target \\
  $e_1$ & Encoding of information desired for predicting $y$ \\
  $e_2$ & Encoding of information \textit{not} desired for predicting $y$ \\
  $\tilde{e}_1$ & Noisy version of $e_1$ used with $e_2$ for reconstructing $x$ \\
  $z$ & Undesired information not to be used for predicting $y$ \\
  \hline
  $f_i$ & An atomic factor of data \\
  $F$ & Set of underlying atomic factors of data $F = \{f_i\}$ \\
  $F_y$ & Subset of $F$ that is informative of $y$ \\
  $\overline{F}_y$ & Subset of $F$ that is \textit{not} informative of $y$ \\
  $F_b$ & Subset of $F_y$ that is biased \\
  \hline
  $Enc$ & Encoder that embeds $x$ into $e = [e_1 \ \ e_2]$ \\
  $Pred$ & Predictor that infers $y$ from $e_1$ \\
  $\psi$ & Noisy transformer that converts $e_1$ to $\tilde{e}_1$, e.g., Dropout \\
  $Dec$ & Decoder that reconstructs $x$ from $[\tilde{e}_1 \ \ e_2]$ \\
  $Dis_1$ & Adversarial disentangler that tries to predict $e_2$ from $e_1$ \\
  $Dis_2$ & Adversarial disentangler that tries to predict $e_1$ from $e_2$ \\
  $D_z$ & Adversarial $z$-discriminator that tries to predict $z$ from $e_1$ \\
  \hbline
\end{tabular}
\end{table}
\setlength{\tabcolsep}{\defaulttabcolsep} 

\subsection{Unified Invariance Induction Framework}
\label{sec:uaif_gen}

Data samples ($x$) can be abstractly decomposed into a set of underlying \emph{atomic} factors of variation $F = \{f_i\}$. This set can be as simple as a collection of numbers denoting the position of a point in space or as complicated as information pertaining to various facial attributes that combine non-trivially to form the image of someone's face. Modeling the interactions between factors of data is an open problem. However, supervised learning of the mapping of $x$ to target ($y$) involves a relatively narrower (yet challenging) problem of finding those factors of variation ($F_y$) that contain all the information required for predicting $y$ and discarding all the others ($\overline{F}_y$). Thus, $F_y$ and $\overline{F}_y$ form a partition of $F$, where we are more interested in the former than the latter. Since $y$ is independent of $\overline{F}_y$, i.e., $y \perp \overline{F}_y$, we get $p(y|x) = p(y|F_y)$. Estimating $p(y|x)$ as $q(y|F_y)$ from data is beneficial because the nuisance factors (i.e., $f_i \perp y$), which comprise $\overline{F}_y$, are never presented to the estimator, thus avoiding inaccurate learning of associations between nuisance factors and $y$.

We incorporate the idea of splitting $F$ into $F_y$ and $\overline{F}_y$ in our framework in a more relaxed sense as learning a split latent representation of $x$ in the form of $e = [e_1 \ e_2]$. While $e_1$ aims to capture all the information relevant for predicting the target ($F_y$), $e_2$ contains nuisance factors ($\overline{F}_y$). Once trained, the model can be used to infer $e_1$ from $x$ followed by $y$ from $e_1$. Learning such a representation of data requires careful separation of information of $x$ into two independent latent embeddings. We bring about this information separation in our framework through competition between the task of predicting $y$ and that of reconstructing $x$, coupled with enforced disentanglement between the two representations. This competition is induced by requiring the model to predict $y$ from $e_1$ while being able to reconstruct $x$ from $e_2$ along with a noisy version of $e_1$. Thus, the prediction task is favored if $e_1$ encodes everything in $x$ that is informative of $y$ while reconstruction benefits from embedding all information of $x$ into $e_2$, but the disentanglement constraint forces $e_1$ and $e_2$ to contain independent information.

More formally, our general framework for invariance to nuisance consists of four core modules: (1) an encoder $Enc$ that embeds $x$ into $e = [e_1 \ e_2]$, (2) a predictor $Pred$ that infers $y$ from $e_1$, (3) a noisy-transformer $\psi$ that converts $e_1$ into its noisy version $\tilde{e}_1$, and (4) a decoder $Dec$ that reconstructs $x$ from $\tilde{e}_1$ and $e_2$. Additionally, the training objective is equipped with a loss that enforces disentanglement between $Enc(x)_1 = e_1$ and $Enc(x)_2 = e_2$. The training objective for this system can be written as Equation~\ref{eq:uif}:
\begin{align}
\label{eq:uif}
L_n = \alpha L_{pred}(y, &Pred(e_1)) + \beta L_{dec}(x, Dec(\psi(e_1),e_2)) \nonumber \\
&+ \gamma L_{dis}((e_1, e_2)) \nonumber \\ 
= \alpha L_{pred}(y, &Pred(Enc(x)_1)) \nonumber \\
+ \beta L&_{dec}(x, Dec(\psi(Enc(x)_1), Enc(x)_2)) \nonumber \\
&+ \gamma L_{dis}(Enc(x))
\end{align}
where $\alpha$, $\beta$, and $\gamma$ are the importance-weights for the corresponding losses. As evident from the formal objective, the predictor and the decoder are designed to enter into a \textit{competition}, where $Pred$ tries to pull information relevant to $y$ into $e_1$ while $Dec$ tries to extract all the information about $x$ into $e_2$. This is made possible by $\psi$, which makes $\tilde{e}_1$ an unreliable source of information for reconstructing $x$. Moreover, a version of this framework without $\psi$ can converge to a degenerate solution where $e_1$ contains all the information about $x$ and $e_2$ contains nothing (noise), because absence of $\psi$ allows $e_1$ to be readily available to $Dec$. The competitive pulling of information into $e_1$ and $e_2$ induces information separation --- $e_1$ tends to contain more information relevant for predicting $y$ and $e_2$ more information irrelevant to the prediction task. However, this competition is not sufficient to completely partition information of $x$ into $e_1$ and $e_2$. Without the disentanglement term ($L_{dis}$) in the objective, $e_1$ and $e_2$ can contain redundant information such that $e_2$ has information relevant to $y$ and, more importantly, $e_1$ contains nuisance factors. The disentanglement term in the training objective encourages the desired clean partition. Thus, \emph{essential} factors required for predicting $y$ concentrate into $e_1$ and all other factors migrate to $e_2$.

\begin{figure*}[t]
\centering
\includegraphics[trim={0 1.05cm 0 0.85cm},clip,width=0.75\textwidth]{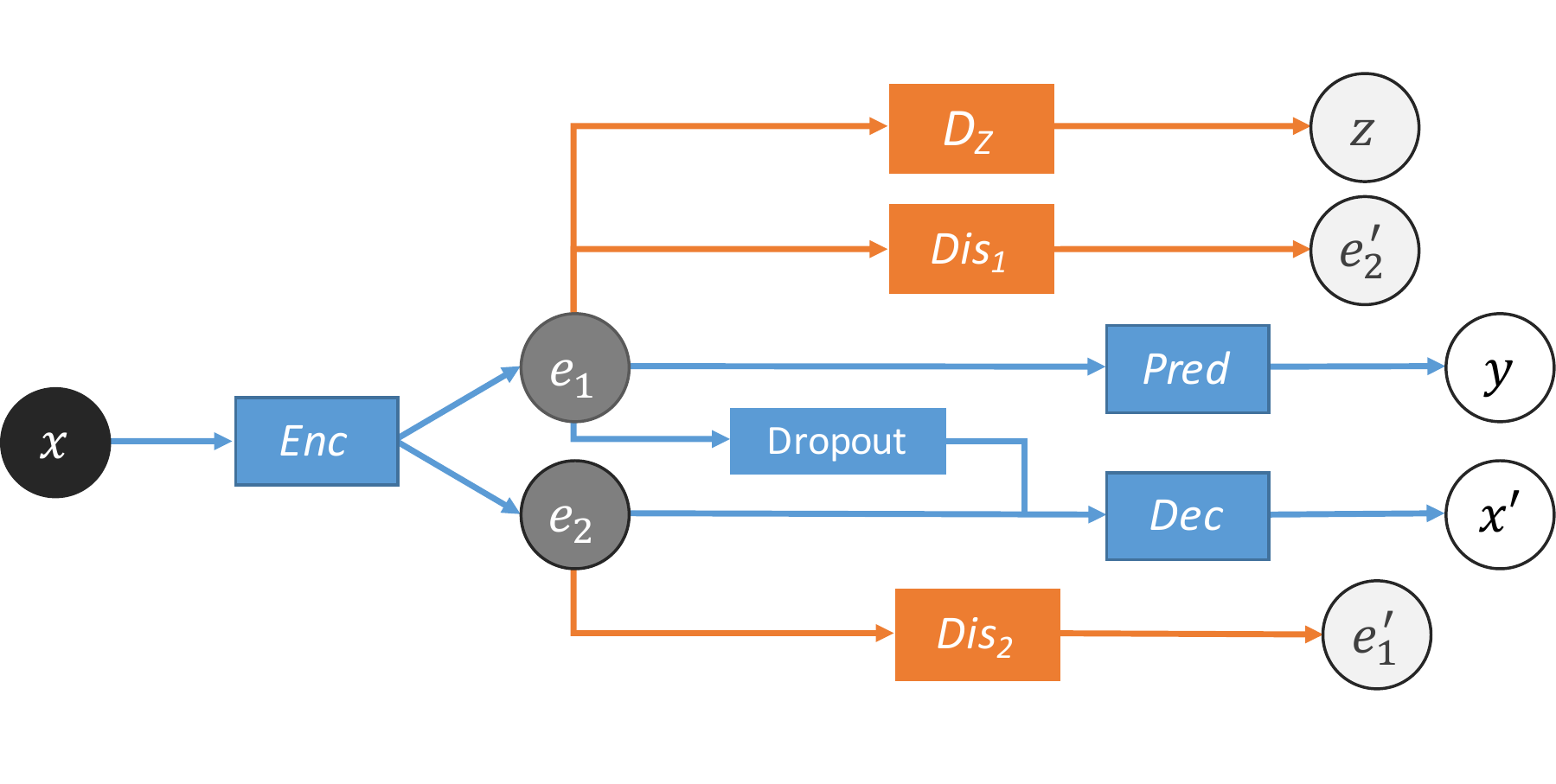}
\caption{\label{fig:hai}The Unified Adversarial Invariance (UnifAI) model. $Enc$ encodes $x$ into $e_1$ and $e_2$. $Pred$ uses $e_1$ to predict $y$. $Dec$ uses $\psi(e_1)$ and $e_2$ to reconstruct $x$. $\psi$ is implemented as dropout. Disentanglement is enforced through adversarial modules $Dis_1$ and $Dis_2$. Biasing factors are eliminated from $e_1$ through $D_z$.}
\end{figure*}

While nuisance factors $\overline{F}_y$ can be separated from those essential for $y$-prediction using the $L_n$ objective in Equation~\ref{eq:uif}, biasing factors cannot. This is because biasing factors are correlated with $y$ and, hence, form a subset $F_b$ of $F_y$, i.e., $F_b \subseteq F_y$. The $L_n$ objective has no way to determine whether an essential factor is biased. In general, this is true for fairness settings. External information about biasing $z$ (encompassing $F_b$) and training mechanisms that use this information to eliminate $z$ from the latent representation are necessary for making fair $y$-predictions, even if it entails relatively poor performance at the task of predicting $y$. In order to achieve this, we augment $L_n$ with a loss term $L_z$ that penalizes $e_1$ for containing $z$ information. The $L_z$ loss can be abstractly viewed as a proxy for the mutual information $I(e_1:z)$. The final training objective is as shown in Equation~\ref{eq:hif}.
\begin{align}
\label{eq:hif}
L &= L_n + \delta L_z(e_1) \nonumber \\ 
&= \alpha L_{pred}(y, Pred(Enc(x)_1)) \nonumber \\
&\qquad+ \beta L_{dec}(x, Dec(\psi(Enc(x)_1), Enc(x)_2)) \nonumber \\
&\qquad\qquad+ \gamma L_{dis}(Enc(x)) + \delta L_z(Enc(x)_1)
\end{align}
The effect of $L_z$ on the training objective is very intuitive. It forces unwanted $z$ out of $e_1$, such that $e_1$ encodes $F_y \setminus F_b$. While $L_z$ is in direct conflict with $L_{pred}$ for biasing $z$, $L_{dec}$ and $L_{dis}$ are not. The decoder can still receive the $z$ forced out of $e_1$ through $e_2$, which encodes $\overline{F}_y \bigcup F_b$, and use them for reconstructing $x$. The $L_{dis}$ loss is unaffected because it only enforces disentanglement between $e_1$ and $e_2$, and removing $z$ from $e_1$ does not violate that.


\subsection{Adversarial Model Design and Optimization}

While there are numerous ways to implement the proposed invariance induction framework, e.g., using mutual information and variational approximation tools similar to~\cite{bib:cvib}, we adopt an adversarial model design, introducing a novel approach to disentanglement in the process. $Enc$, $Pred$ and $Dec$ are modeled as neural networks. $\psi$ can be modeled as a parametric noisy-channel, where the parameters of $\psi$ can also be learned during training. However, we model $\psi$ as multiplicative Bernoulli noise using dropout~\cite{bib:dropout} since it provides a straightforward method for noisy-transformation of $e_1$ into $\tilde{e}_1$ without complicating the training process.

We augment these core modules with two adversarial \textit{disentanglers} -- $Dis_1$ and $Dis_2$. While $Dis_1$ aims to predict $e_2$ from $e_1$, $Dis_2$ aims to do the inverse. It would be impossible to predict either embedding from the other if they were truly independent. Hence, the objectives of the two disentanglers are in direct opposition to the desired disentanglement, forming the basis for adversarial minimax optimization. In comparison to the use of information theoretic measures like the mutual information $I(e_1:e_2)$ (or a suitable proxy) for the loss $L_{dis}$, this approach to disentanglement does not require $e_1$ and $e_2$ to be stochastic, and does not assume prior distributions for the two embeddings.

Thus, $Enc$, $Pred$ and $Dec$ can be thought of as a composite model ($M_1$) that is pitted against another composite model ($M_2$) containing $Dis_1$ and $Dis_2$. This results in an adversarial instantiation of the framework for invariance to nuisance factors. In order to complete the adversarial model so that it allows removal of known $z$ from $e_1$, a $z$-discriminator $D_z$ is added to the model that aims to predict $z$ from $e_1$. Thus, the objective of $D_z$ is the opposite of the desired invariance to $z$, making it a natural component of the composite model $M_2$ for fairness settings.

Figure~\ref{fig:hai} shows our unified adversarial invariance (UnifAI) model for invariance to nuisance as well as biasing factors. The composite model $M_1$ is represented by the color blue and $M_2$ with orange. The model is trained through backpropagation by playing a minimax game. The objective for invariance to nuisance factors is shown in Equation~\ref{eq:uai}.
\begin{align}
&\min_{Enc, Pred, Dec} \ \ \max_{Dis_1, Dis_2} J_n \ ; \ \text{where:} \nonumber \\
J_n&(Enc, Pred, Dec, Dis_1, Dis_2) \nonumber \\
&= \ \alpha L_{pred} \bigl(y, Pred(e_1) \bigr) + \beta L_{dec} \bigl(x, Dec(\psi(e_1), e_2)\bigr) \nonumber \\
&\qquad\qquad\qquad+ \gamma \tilde{L}_{dis} \bigl((e_1, e_2) \bigr) \nonumber \\
&= \ \alpha L_{pred} \bigl(y, Pred(Enc(x)_1) \bigr) \nonumber \\
&\qquad+ \beta L_{dec} \bigl(x, Dec(\psi(Enc(x)_1)), Enc(x)_2))\bigr) \nonumber \\
&\qquad\qquad+ \ \gamma \bigl\{\tilde{L}_{dis_1} \bigl(Enc(x)_2, Dis_1(Enc(x)_1)\bigr) \nonumber \\
&\qquad\qquad\qquad+ \tilde{L}_{dis_2} \bigl(Enc(x)_1, Dis_2(Enc(x)_2)\bigr)\bigr\}
\label{eq:uai}
\end{align}
Equation~\ref{eq:hai} describes the complete minimax objective for invariance to both nuisance and biasing factors of data.
\begin{align}
&\min_{Enc, Pred, Dec} \ \ \max_{Dis_1, Dis_2, D_z} J \ ; \ \text{where:} \nonumber \\
J&(Enc, Pred, Dec, Dis_1, Dis_2, D_z) \nonumber \\
&= \ J_n(Enc, Pred, Dec, Dis_1, Dis_2) + \delta \tilde{L}_z \bigl(z, D_z(e_1)\bigr) \nonumber \\
&= \ \alpha L_{pred} \bigl(y, Pred(Enc(x)_1) \bigr) \nonumber \\
&\qquad+ \beta L_{dec} \bigl(x, Dec(\psi(Enc(x)_1)), Enc(x)_2))\bigr) \nonumber \\
&\qquad\qquad+ \ \gamma \bigl\{\tilde{L}_{dis_1} \bigl(Enc(x)_2, Dis_1(Enc(x)_1)\bigr) \nonumber \\
&\qquad\qquad\qquad+ \tilde{L}_{dis_2} \bigl(Enc(x)_1, Dis_2(Enc(x)_2)\bigr)\bigr\} \nonumber \\
&\qquad\qquad\qquad\qquad+ \delta \tilde{L}_z \bigl(z, D_z(Enc(x)_1)\bigr)
\label{eq:hai}
\end{align}
We optimize the proposed adversarial model using a scheduled update scheme where we freeze the weights of a composite player model ($M_1$ or $M_2$) when we update the weights of the other. $M_2$ should ideally be trained to convergence before updating $M_1$ in each training epoch to backpropagate accurate and stable disentanglement-inducing and $z$-eliminating gradients to $Enc$. However, this is not scalable in practice. We update $M_1$ and $M_2$ with a frequency of $1:k$. We found $k = 5$ to perform well in our experiments, but a larger $k$ might be required depending on the complexity of the prediction task, the unwanted variables, and the dataset in general. We use mean squared error for the disentanglement losses $\tilde{L}_{dis_1}$ and $\tilde{L}_{dis_2}$. The discriminative loss $\tilde{L}_z$ depends on the nature of $z$-annotations, e.g., cross-entropy loss for categorical $z$.

Adversarial training with $Dis_1$, $Dis_2$, and $D_z$ necessitates the choice of appropriate adversarial targets, i.e., the targets that are used to calculate losses and gradients from the adversaries to update the encoder. More specifically, in the $M_2$ phase of the scheduled training, at a given iteration, the targets for calculating $\tilde{L}_{dis_1}$ and $\tilde{L}_{dis_2}$ are the true values of the vectors $e_2$ and $e_1$, respectively, calculated from $x$ at that iteration. On the other hand, in the $M_1$ phase, the targets for $\tilde{L}_{dis_1}$ and $\tilde{L}_{dis_2}$ are randomly sampled vectors. The intuition behind this choice of targets is straightforward -- for truly disentangled $e^*_1$ and $e^*_2$, the best an adversary predicting one from the other can do is predict random noise because $e^*_1 \perp e^*_2$ and their mutual information is zero. Hence, the encoder should be updated in a way that the best these disentanglers can do is predict random noise. We implement this by constraining the encoder to use the hyperbolic tangent activation in its final layer, thus limiting the components of $e_1$ and $e_2$ to $[-1, 1]$ (any other bounded activation function could be used), and sampling random vectors from a uniform distribution in $[-1, 1]$ as targets for the $M_1$ phase. Similarly, for biasing factors, ground-truth $z$ is used as the target in $M_2$ phase for $\tilde{L}_z$ while random $z$ are used as targets in the $M_1$ phase. For categorical $z$, this is implemented as a straightforward sampling of $z$ from the empirically estimated categorical distribution of $z$ calculated from the training dataset.

\subsection{Invariant Predictions with the Trained Model}

The only components of the proposed framework that are required for making predictions at test time are the encoder $Enc$ and the predictor $Pred$. Prediction is a simple forward-pass of the graph $x \dashedrightarrow e_1 \dashedrightarrow y$. Thus, making predictions with a model trained in the proposed framework does not have any overhead computational cost.


\section{Analysis}
\label{sec:theoretical}

We analyze the relationship between the loss weights $\alpha$ and $\beta$, corresponding to the competing tasks of predicting $y$ and reconstructing $x$, respectively, in our generalized invariance induction framework. We then discuss the equilibrium of the minimax game in our adversarial instantiation for both nuisance and biasing factors. Finally, we use the results of these two analyses to provide a systematic way for tuning the loss weights $\alpha$ $\beta$, and $\gamma$. The following analyses are conducted assuming a model with infinite capacity, i.e., in a non-parametric limit.

\paragraph*{Competition between prediction and reconstruction} The prediction and reconstruction tasks in our framework are designed to compete with each other for invariance to nuisance factors. Thus, $\eta = \frac{\alpha}{\beta}$ influences which task has higher priority in the objective shown in Equation~\ref{eq:uif}. We analyze the affect of $\eta$ on the behavior of our framework at optimality considering perfect disentanglement of $e_1$ and $e_2$. There are two asymptotic scenarios with respect to $\eta$ -- (1) $\eta \rightarrow \infty$ and (2) $\eta \rightarrow 0$. In case (1), our framework for invariance to nuisance (i.e., without $D_z$) reduces to a predictor model, where the reconstruction task is completely disregarded ($\beta \ll \alpha$). Only the branch $x \dashedrightarrow e_1 \dashedrightarrow y$ remains functional. Consequently, $e_1$ contains all $f \in F'$ at optimality, where $F_y \subseteq F' \subseteq F$. In contrast, case (2) reduces the framework to an autoencoder, where the prediction task is completely disregarded ($\beta \gg \alpha$), and only the branch $x \dashedrightarrow e_2 \dashedrightarrow x'$ remains functional because the other input to $Dec$, $\psi(e_1)$, is noisy. Thus, $e_2$ contains all $f \in F$ and $e_1$ contains nothing at optimality, under perfect disentanglement. In transition from case (1) to case (2), by keeping $\alpha$ fixed and increasing $\beta$, the reconstruction loss starts contributing more to the overall objective, thus inducing more competition between the two tasks. As $\beta$ is gradually increased, $f \in (F' \smallsetminus F_y) \subseteq \overline{F}_y$ migrate from $e_1$ to $e_2$ because $f \in \overline{F}_y$ are irrelevant to the prediction task but can improve reconstruction by being more readily available to $Dec$ through $e_2$ instead of $\psi(e_1)$. After a point, further increasing $\beta$ is, however, detrimental to the prediction task as the reconstruction task starts dominating the overall objective and pulling $f \in F_y$ from $e_1$ to $e_2$. Results in Section~\ref{subsec:eta} show that this intuitive analysis is consistent with the observed behavior.

In the case of known undesired $z$, the presence of $D_z$ in the unified framework pushes known $z$ out of $e_1$, thus favoring the reconstruction objective by forcing known $z$ to migrate to $e_2$. Thus, the analysis of the competition still holds intuitively for nuisance factors besides $z$.

\setlength{\tabcolsep}{1.5em} 
\begin{table*}
\makegapedcells
\centering
\caption{\label{tab:eyb}Results on Extended Yale-B dataset. High $A_y$ and low $A_z$ are desired.}
\begin{tabular}{ c ?{1.5pt} c ?{0.75pt} c ?{0.75pt} c ?{0.75pt} c ?{1.5pt} c ?{0.25pt} c ?{0.25pt} c }
  \hbline
  \textbf{Metric} & \textbf{NN+MMD~\cite{bib:nnmmd}} & \textbf{VFAE~\cite{bib:vfae}} & \textbf{CAI~\cite{bib:sai}} & \textbf{CVIB~\cite{bib:cvib}} & \textbf{UnifAI (ours)} & $\boldsymbol{B_1}$ & $\boldsymbol{B_0}$ \\
  \hbline
  $A_y$ & 0.82 & 0.85 & 0.89 & 0.82 & \textbf{0.95} & 0.94 & 0.90 \\
  $A_z$ & - & 0.57 & 0.57 & 0.45 & \textbf{0.24} & 0.28 & 0.60 \\
  \hbline
\end{tabular}
\end{table*}
\setlength{\tabcolsep}{\defaulttabcolsep} 

{
\def \fs {0.4}
\def \sfs {0.81}
\begin{figure*}
\centering
\begin{subfigure}{\fs\textwidth}
\centering
\includegraphics[width=\sfs\textwidth]{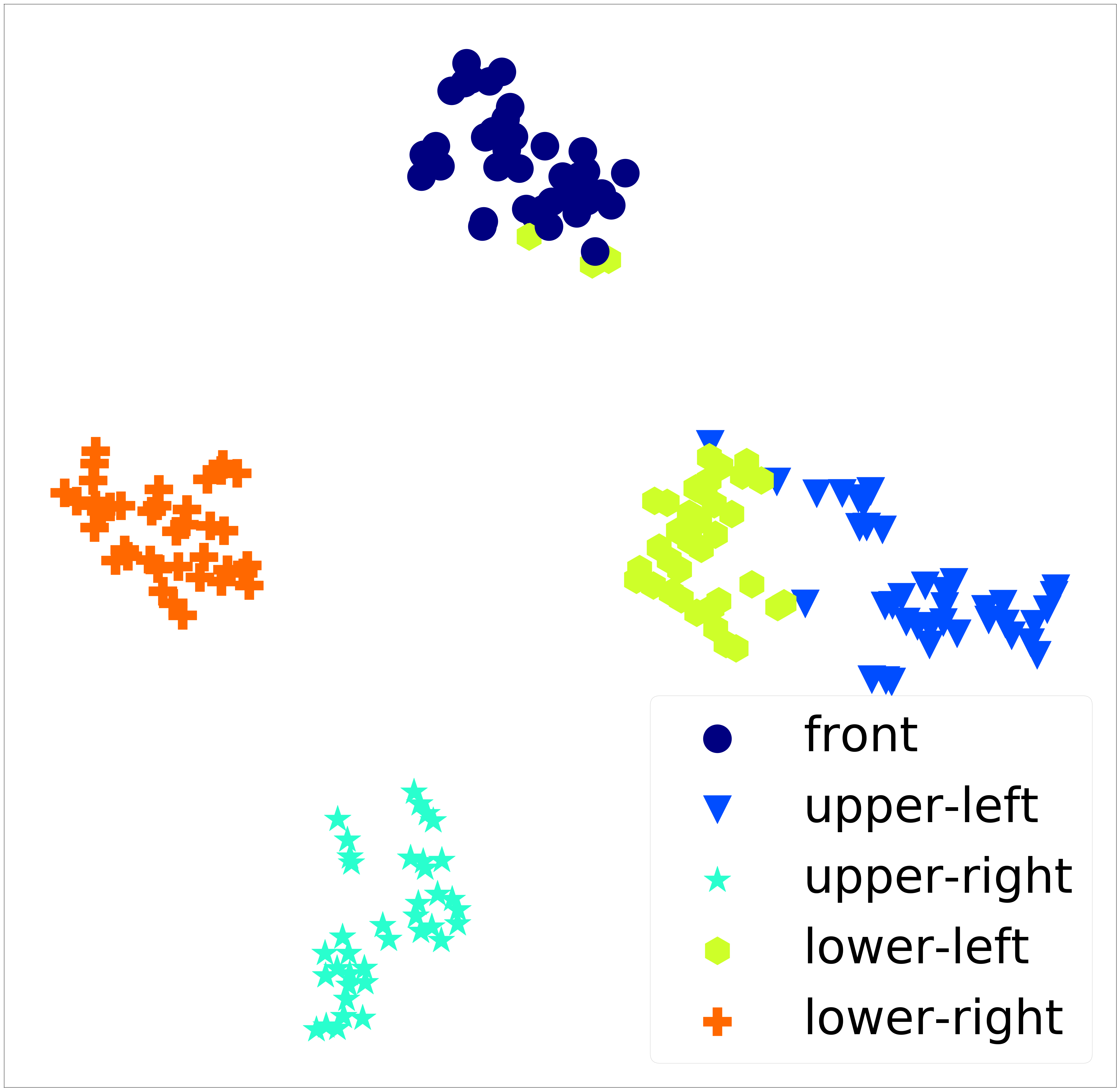}
\caption{}
\end{subfigure}
\begin{subfigure}{\fs\textwidth}
\centering
\includegraphics[width=\sfs\textwidth]{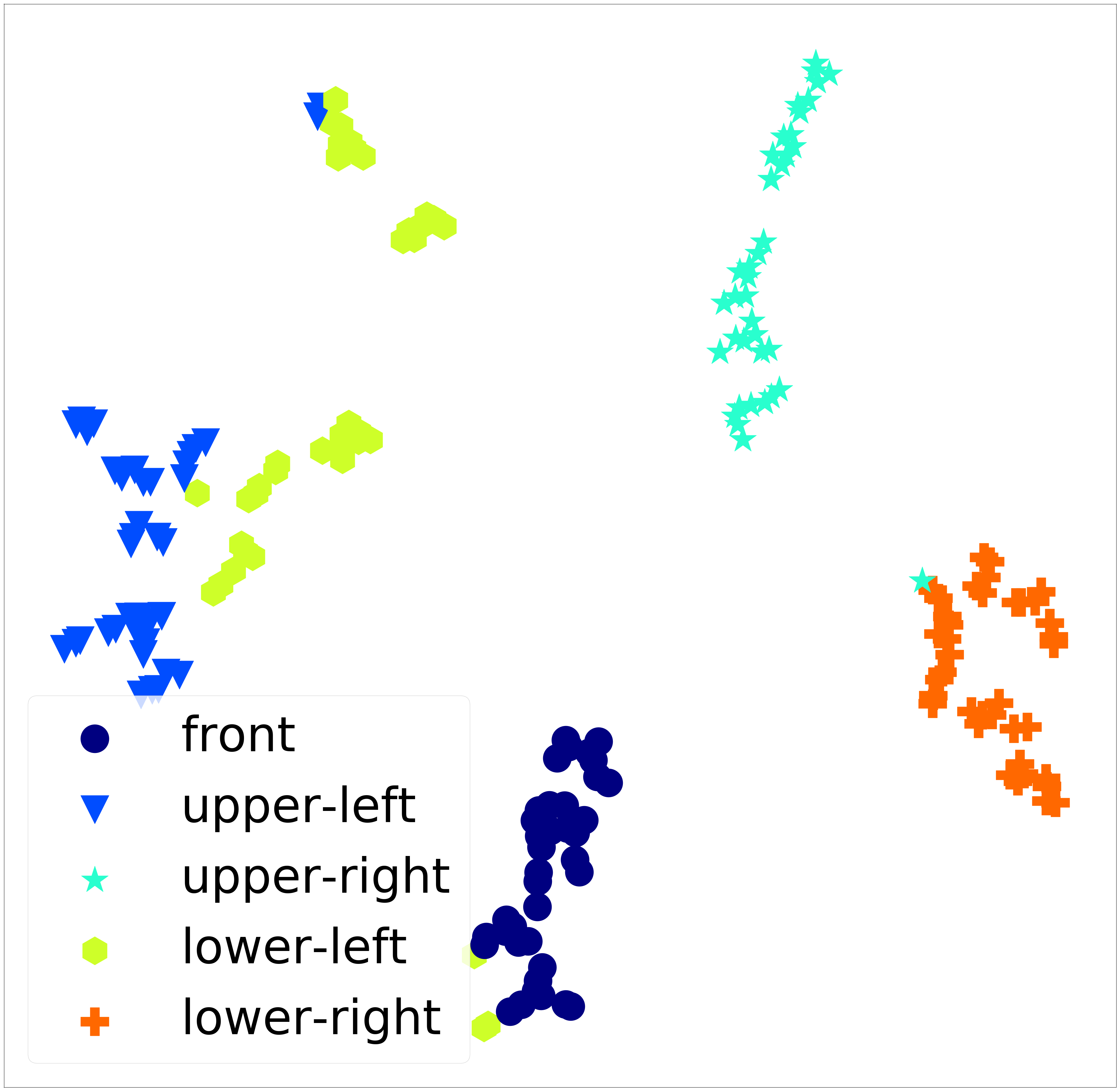}
\caption{}
\end{subfigure}
\begin{subfigure}{\fs\textwidth}
\centering
\includegraphics[width=\sfs\textwidth]{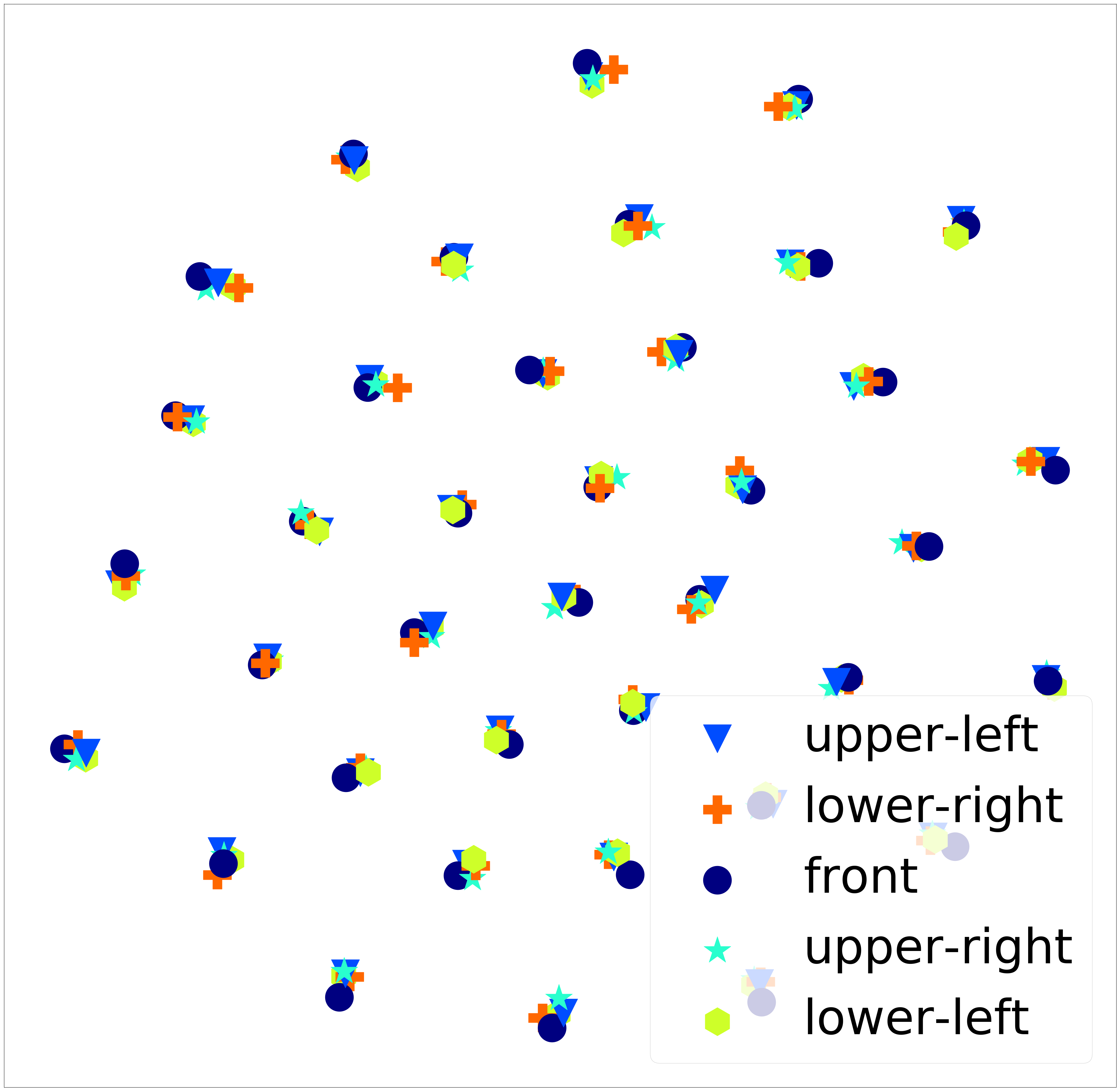}
\caption{}
\end{subfigure}
\begin{subfigure}{\fs\textwidth}
\centering
\includegraphics[width=\sfs\textwidth]{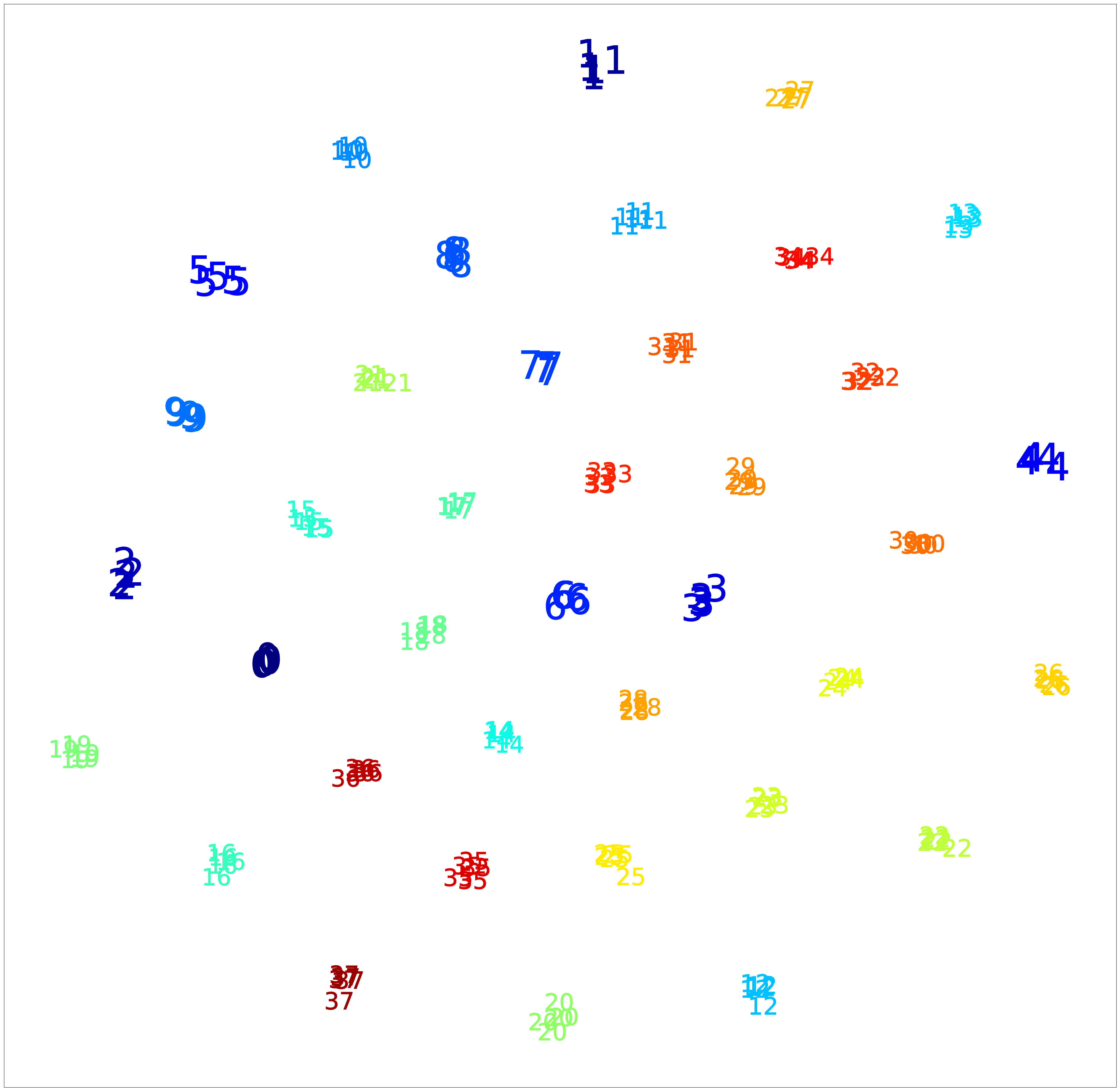}
\caption{}
\end{subfigure}
\caption{Extended Yale-B -- t-SNE visualization of (a) raw data, (b) $e_2$ labeled by lighting condition, (c) $e_1$ labeled by lighting condition, and (d) $e_1$ labeled by subject-ID (numerical markers, not colors). Raw images cluster by lighting. $e_1$ clusters by identity but not lighting, as desired, while $e_2$ clusters by lighting.}
\label{fig:tsne_eyb}
\end{figure*}
}

\paragraph*{Equilibrium analysis of adversarial instantiation} The disentanglement and prediction objectives in our adversarial model design can simultaneously reach an optimum where $e_1$ contains $F_y$ and $e_2$ contains $\overline{F}_y$. Hence, the minimax objective in our method has a \emph{win-win equilibrium} for invariance to nuisance factors. However, the training objective of $D_z$ for biasing $z$ is in direct opposition to the prediction task because such $z$ are correlated with $y$. This leads to a win-lose equilibrium for biasing factors, which is true in general for all methods of fair representation learning.

\paragraph*{Selecting loss weights} Using the above analyses, any $\gamma$ that successfully disentangles $e_1$ and $e_2$ should be sufficient. We found $\gamma = 1$ to work well for the datasets on which we evaluated the proposed model. On the other hand, if $\gamma$ is fixed, $\alpha$ and $\beta$ can be selected by starting with $\beta \ll \alpha$ and gradually increasing $\beta$ as long as the performance of the prediction task improves. The removal of biasing $z$ is controlled by the loss weight $\delta$ in Equation~\ref{eq:hai} and requires $\delta$ to be carefully tuned depending on the complexity of the dataset, the prediction task, and the biasing factors.



\section{Empirical Evaluation}
\label{sec:evaluation}

\begin{figure*}
\centering
\includegraphics[width=0.75\textwidth]{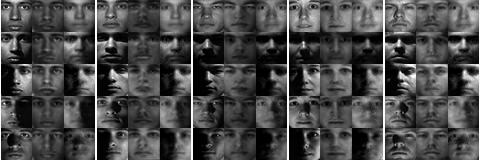}
\caption{\label{fig:recon_eyb}Extended Yale-B -- reconstruction results. Each block shows results for a single subject. Columns in each block are (left to right): real image, reconstruction from $e_1$ and that from $e_2$. Reconstructions from $e_1$ show that it captures subject-identity but has little lighting information, thus achieving the invariance goal. Reconstructions from $e_2$ show that it captures lighting but not identity. Viewing along rows across blocks, it is easy to see that reconstructions from $e_2$ look similar.}
\end{figure*}

\setlength{\tabcolsep}{1.5em} 
\begin{table*}
\makegapedcells
\centering
\caption{\label{tab:chairs}Results on Chairs. High $A_y$ and low $A_z$ are desired.}
\begin{tabular}{ c ?{1.5pt} c ?{0.75pt} c ?{0.75pt} c ?{0.75pt} c ?{1.5pt} c ?{0.25pt} c ?{0.25pt} c }
  \hbline
  \textbf{Metric} & \textbf{NN+MMD~\cite{bib:nnmmd}} & \textbf{VFAE~\cite{bib:vfae}} & \textbf{CAI~\cite{bib:sai}} & \textbf{CVIB~\cite{bib:cvib}} & \textbf{UnifAI (ours)} & $\boldsymbol{B_1}$ & $\boldsymbol{B_0}$ \\
  \hbline
  $A_y$ & 0.70 & 0.72 & 0.68 & 0.67 & \textbf{0.74} & 0.69 & 0.67 \\
  $A_z$ & 0.43 & 0.37 & 0.69 & 0.52 & \textbf{0.34} & 0.54 & 0.70 \\
  \hbline
\end{tabular}
\end{table*}
\setlength{\tabcolsep}{\defaulttabcolsep} 

We provide empirical results on five tasks relevant to invariant feature learning for robustness to nuisance and fair predictions: (1) invariance to inherent nuisance factors, (2) effective use of synthetic data augmentation for learning invariance to specific nuisance factors, (3) learning invariance to \emph{arbitrary} nuisance factors by leveraging Generative Adversarial Networks, (4) domain adaptation through learning invariance to ``domain'' information, and (5) fair representation learning. For experiments (1)--(4), we do not use nuisance annotations for learning invariance, i.e., we train the model without $D_z$. In contrast, the state-of-the-art methods use $z$-labels. We evaluate the performance of our model and prior works on two metrics -- accuracy of predicting $y$ from $e_1$ ($A_y$) and accuracy of predicting $z$ from $e_1$ ($A_z$). While $A_y$ is calculated directly from the predictions of the trained models, $A_z$ is calculated using a two-layer neural network trained \textit{post hoc} to predict $z$ from the latent embedding. The goal of the model is to achieve high $A_y$ in all cases but $A_z$ close to random chance for nuisance factors and $A_z$ the same as the population share of the majority $z$-class for biasing factors in fairness settings. We train two baseline versions of our model for our ablation experiments --- $B_0$ composed of $Enc$ and $Pred$, i.e., a single feed-forward network $x \dashedrightarrow h \dashedrightarrow y$ and $B_1$, which is the same as the composite model $M_1$, i.e., the proposed model trained without the adversarial components. $B_0$ is used to validate the phenomenon that invariance to nuisance by exclusion is a better approach than robustness through inclusion whereas $B_1$ helps evaluate the importance of disentanglement. Hence, results of $B_0$ and $B_1$ are presented for tasks (1) and (2). Besides the results on the aforementioned tasks, we provide empirical insight into the competition between the prediction and reconstruction tasks in our framework, as discussed in Section~\ref{sec:theoretical}, through the influence of the ratio $\frac{\alpha}{\beta}$ on $A_y$.

{
\def \fs {0.32}
\def \sfs {0.9}
\begin{figure*}[h]
\centering
\begin{subfigure}{\fs\textwidth}
\centering
\includegraphics[width=\sfs\textwidth]{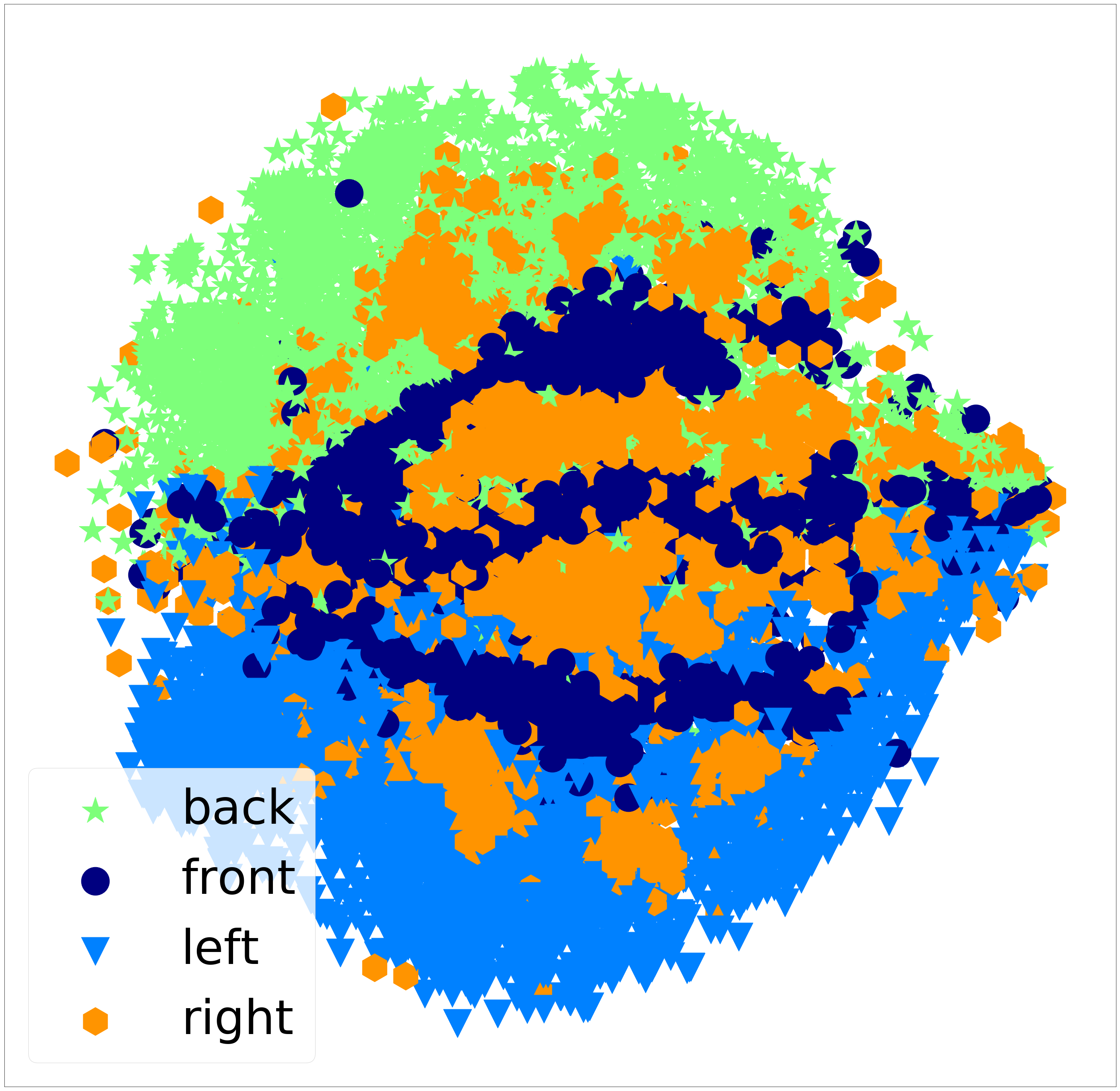}
\caption{Raw data}
\end{subfigure}
\begin{subfigure}{\fs\textwidth}
\centering
\includegraphics[width=\sfs\textwidth]{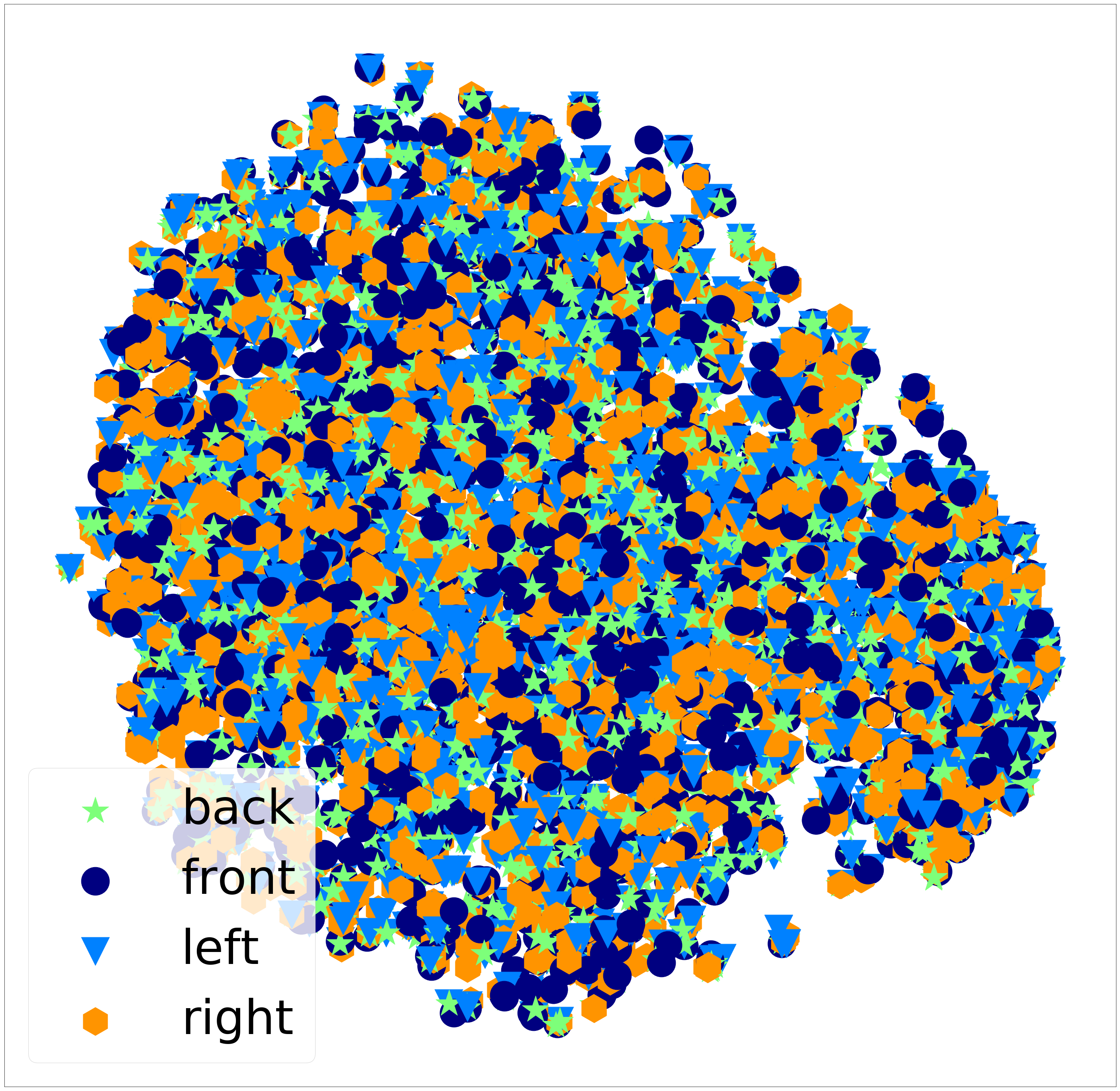}
\caption{$e_1$ embedding}
\end{subfigure}
\begin{subfigure}{\fs\textwidth}
\centering
\includegraphics[width=\sfs\textwidth]{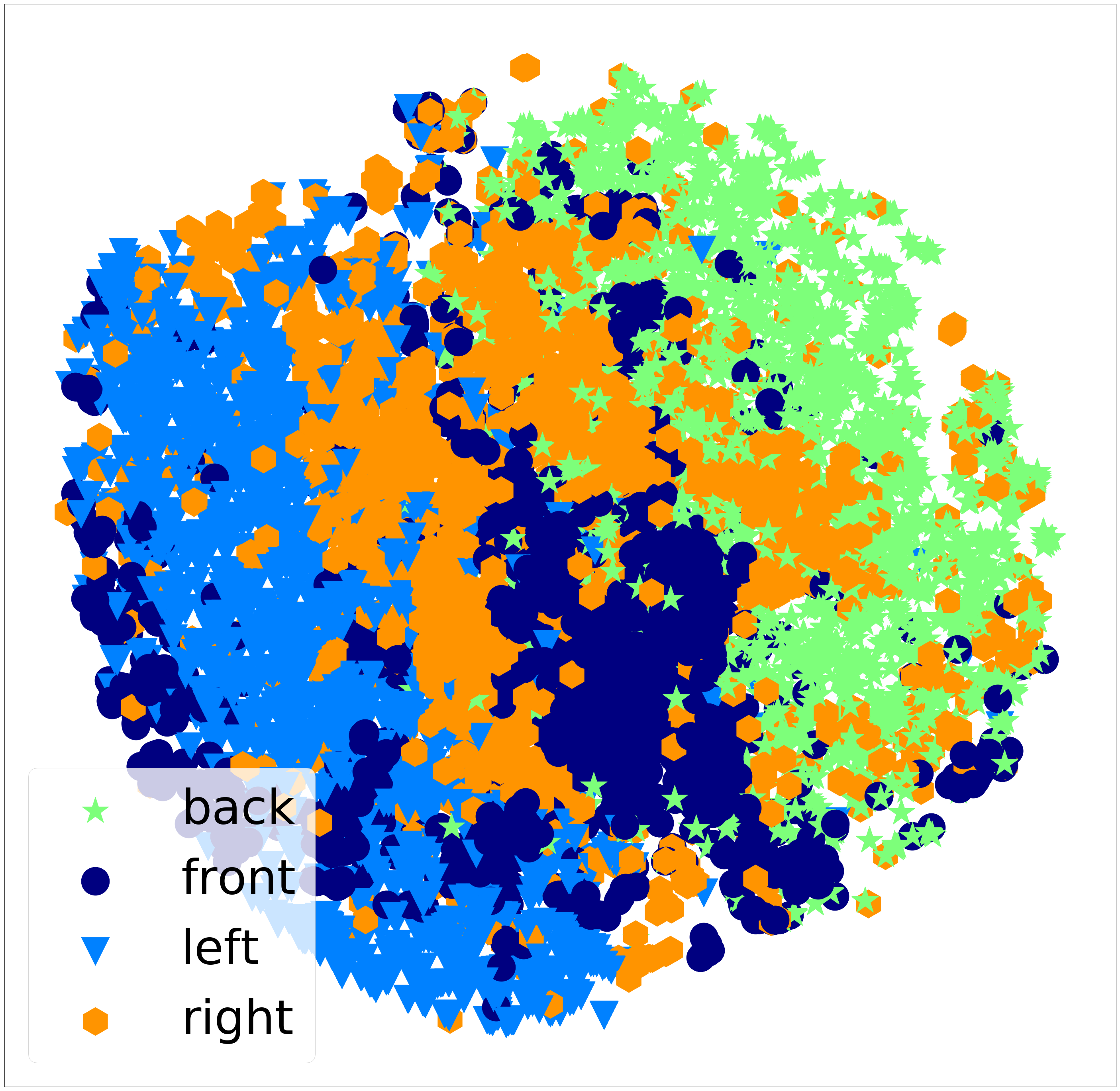}
\caption{$e_2$ embedding}
\end{subfigure}
\caption{\label{fig:tsne_chairs}Chairs dataset -- t-SNE visualization. Labels indicate the nuisance factor -- orientation. Raw images cluster by orientation. $e_1$ clusters by chair-class but not orientation, as desired, while $e_2$ clusters by orientation.}
\end{figure*}
}

\begin{figure*}
\centering
\includegraphics[width=0.75\textwidth]{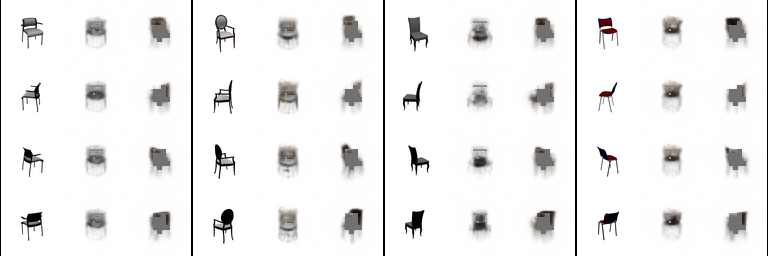}
\caption{\label{fig:recon_chairs}Chairs -- reconstruction results. Each block shows results for a single chair-class. Columns in each block reflect (left to right): real, reconstruction from $e_1$ and that from $e_2$. Reconstructions from $e_1$ show that it captures chair-class but has little orientation information, as desired. Reconstructions from $e_2$ show that it captures orientation but not much about identity.}
\end{figure*}

\subsection{Invariance to inherent nuisance factors}

We provide results of our framework at the task of learning invariance to inherent nuisance factors on two datasets --- Extended Yale-B~\cite{bib:eyb}, which has been used by previous works(\citem{bib:nnmmd,bib:vfae,bib:sai}), and Chairs~\cite{bib:chairs}, which we propose as a new dataset for this task. We compare our framework to existing state-of-the-art invariance induction methods --- CAI~\cite{bib:sai}, VFAE~\cite{bib:vfae}, NN+MMD~\cite{bib:nnmmd}, and CVIB~\cite{bib:cvib}.

\paragraph*{Extended Yale-B} This dataset contains face-images of $38$ subjects under various lighting conditions. The target $y$ is the subject identity whereas the inherent nuisance factor $z$ is the lighting condition. We use the prior works' version of the dataset, which has lighting conditions classified into five groups -- front, upper-left, upper-right, lower-left and lower-right, with the same split as $38 \times 5 = 190$ samples used for training and the rest used for testing (\citem{bib:nnmmd,bib:vfae,bib:sai}). We use the same architecture for the predictor and the encoder as CAI (as presented in~\cite{bib:sai}), i.e., single-layer neural networks, except that our encoder produces two encodings instead of one. We also model the decoder and the disentanglers as single-layer neural networks.

Table~\ref{tab:eyb} summarizes the results. The proposed unsupervised method (trained without $D_z$) outperforms ablation versions of our model and existing state-of-the-art (supervised) invariance induction methods on both $A_y$ and $A_z$, providing a significant boost on $A_y$ and nearly complete removal of lighting information from $e_1$ reflected by $A_z$. Furthermore, the accuracy of predicting $z$ from $e_2$ is $0.89$, which validates its automatic migration to $e_2$. Figure~\ref{fig:tsne_eyb} shows t-SNE~\cite{bib:tsne} visualization of raw data and embeddings $e_1$ and $e_2$ for our model. While raw data is clustered by lighting conditions $z$, $e_1$ exhibits clustering by $y$ with no grouping based on $z$, and $e_2$ exhibits near-perfect clustering by $z$. Figure~\ref{fig:recon_eyb} shows reconstructions from $e_1$ and $e_2$. Dedicated decoder networks were trained (with weights of $Enc$ frozen) to generate these visualizations. As evident, $e_1$ captures identity-related information but not lighting while $e_2$ captures the inverse.

\paragraph*{Chairs} This dataset consists of 1,393 different chair types rendered at $31$ yaw angles and two pitch angles using a computer aided design model. We treat the chair identity as the target $y$ and the yaw angle $\theta$ as the nuisance factor $z$ by grouping $\theta$ into four categories -- front, left, right and back. This $z$ information is used for training previous works but our model is trained without $D_z$ and hence, without any $z$-information. We split the data into training and testing sets by picking alternate yaw angles. Therefore, \emph{there is no overlap of $\theta$ between the two sets}. We model the encoder and the predictor as two-layer neural networks for the previous works and our model. We also model the decoder as a two-layer network and the disentanglers as single-layer networks.

Table~\ref{tab:chairs} summarizes the results, showing that our model outperforms both ablation baselines and previous state-of-the-art methods on both $A_y$ and $A_z$. Moreover, the accuracy of predicting $\theta$ from $e_2$ is $0.73$, which shows that this information migrates to $e_2$. Figure~\ref{fig:tsne_chairs} shows t-SNE visualization of raw data and embeddings $e_1$ and $e_2$ for our model. While raw data and $e_2$ are clustered by the orientation direction $z$, $e_1$ exhibits no grouping based on $z$. Figure~\ref{fig:recon_chairs} shows results of reconstructing $x$ from $e_1$ and $e_2$ generated in the same way as for Extended Yale-B above. The figure shows that $e_1$ contains identity information but nothing about $\theta$ while $e_2$ contains $\theta$ with limited identity information.

\setlength{\tabcolsep}{1.25em} 
\begin{table*}
\makegapedcells
\centering
\caption{\label{tab:mnist_rot}Results on MNIST-ROT. $\Theta = \{0, \pm 22.5^{\degree}, \pm 45^{\degree}\}$ was used for training. High $A_y$ and low $A_z$ are desired. VFAE does not allow for out-of-domain $z$ because the VFAE encoder requires $z$ as input, and $z$ is categorical here.}
\begin{tabular}{ c ?{1.5pt} c ?{1.5pt} c ?{0.75pt} c ?{0.75pt} c ?{0.75pt} c ?{1.5pt} c ?{0.25pt} c ?{0.25pt} c }
  \hbline
  \textbf{Metric} & \textbf{Angle} & \textbf{NN+MMD~\cite{bib:nnmmd}} & \textbf{VFAE~\cite{bib:vfae}} & \textbf{CAI~\cite{bib:sai}} & \textbf{CVIB~\cite{bib:cvib}} & \textbf{UnifAI (ours)} & $\boldsymbol{B_1}$ & $\boldsymbol{B_0}$ \\
  \hbline
  \multirow{3}*{$A_y$}& $\Theta$ & 0.970 & 0.951 & 0.958 & 0.960 & \textbf{0.977} & 0.972 & 0.974 \\
  & $\pm 55^{\degree}$ & 0.831 & - & 0.829 & 0.819 & \textbf{0.856} & 0.829 & 0.826 \\
  & $\pm 65^{\degree}$ & 0.665 & - & 0.663 & 0.674 & \textbf{0.696} & 0.682 & 0.674 \\
  \hbline
  $A_z$ & - & 0.531 & 0.468 & 0.384 & 0.428 & \textbf{0.338} & 0.409 & 0.586 \\
  \hbline
\end{tabular}
\end{table*}

\begin{figure*}
\centering
\includegraphics[width=0.98\textwidth]{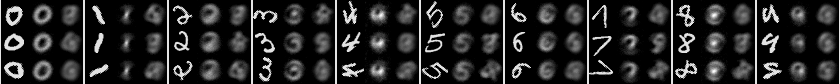}
\caption{MNIST-ROT -- reconstruction results. Each block shows a digit-class. Columns in each block are (left to right): real images, reconstruction from $e_1$ and that from $e_2$. Reconstructions from $e_1$ show that it captures digit-class but has little rotation information, as desired for invariance. Reconstructions from $e_2$ show that it captures rotation as well as other inherent nuisance factors in MNIST digits, which are hard to visually interpret.}
\label{fig:recon_mnist_rot}
\end{figure*}

\begin{figure*}
\centering
\begin{subfigure}{0.4\textwidth}
\centering
\includegraphics[width=0.725\textwidth]{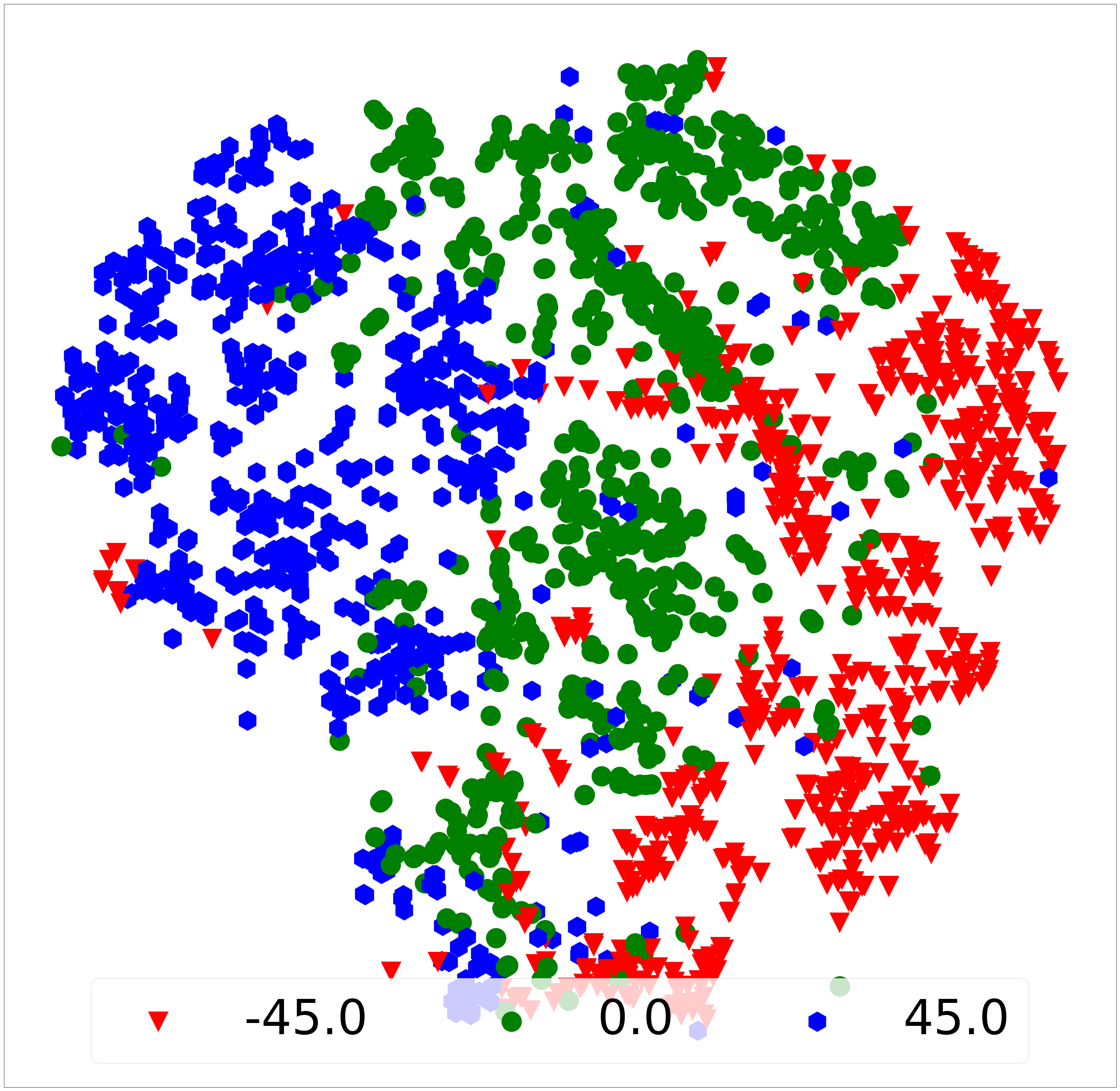}
\caption{Raw data}
\end{subfigure}
\begin{subfigure}{.4\textwidth}
\centering
\includegraphics[width=0.725\textwidth]{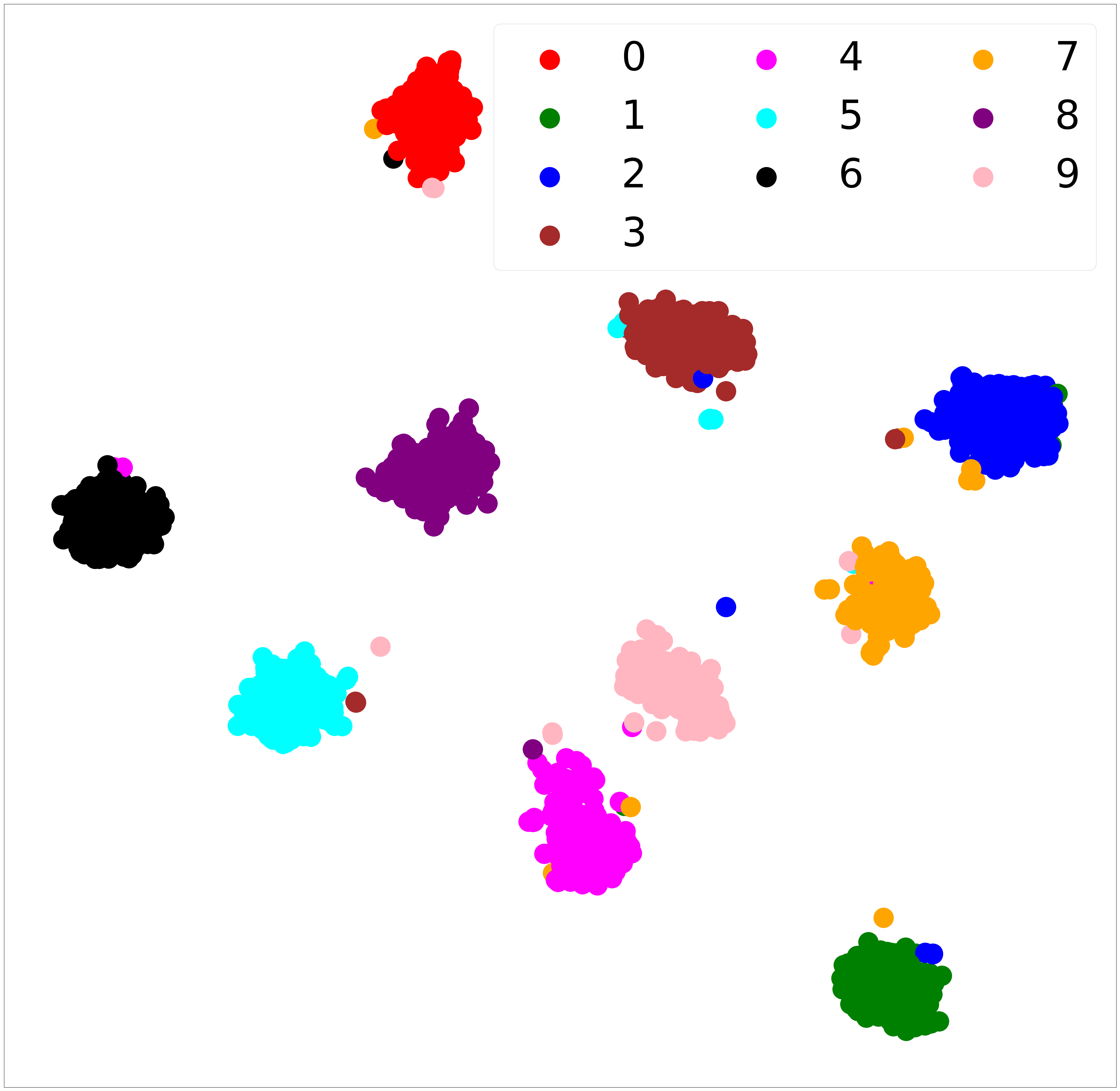}
\caption{$e_1$ embedding}
\end{subfigure}
\caption{MNIST-ROT -- t-SNE visualization. While raw data is clustered by rotation angle $\theta$, $e_1$ is grouped by digit-class.}
\label{fig:tsne_mnist_rot}
\end{figure*}

\subsection{Effective use of synthetic data augmentation for learning invariance}
\label{subsec:data_aug}

Data is often not available for all possible variations of nuisance factors. A popular approach to learn models robust to such expected yet unobserved or infrequently seen (during training) variations is data augmentation through synthetic generation using methods ranging from simple operations~\cite{bib:data_aug_1} like rotation and translation to complex transformations~\cite{bib:face_rec} for synthesis of more sophisticated variations. The prediction model is then trained on the expanded dataset. The resulting model, thus, becomes robust to specific forms of variations of certain nuisance factors that it has seen during training. Invariance induction, on the other hand, aims to completely prevent prediction models from using information about nuisance factors. Data augmentation methods can be more effectively used for improving the prediction of $y$ by using the expanded dataset for inducing invariance by \emph{exclusion} rather than \emph{inclusion}. We use two variants of the MNIST~\cite{bib:mnist} dataset of handwritten digits for experiments on this task. We use the same two-layer architectures for the encoder and the predictor in our model as well as previous works, except that our encoder generates two encodings instead of one. We model the decoder as a three-layer neural network and the disentanglers as single-layer neural networks.

\paragraph*{MNIST-ROT} We create this variant of the MNIST dataset by rotating each image by angles $\theta \in \{-45^{\degree}, -22.5^{\degree}, 0^{\degree}, 22.5^{\degree}, 45^{\degree}\}$ about the Y-axis. We denote this set of angles as $\Theta$. The angle information is used as a one-hot encoding while training the previous works whereas our model is trained without $z$-labels (i.e., without $D_z$). We evaluate all the models on the same metrics $A_y$ and $A_z$ we previously used. We additionally test all the models on $\theta \not \in \Theta$ to gauge the performance of these models on unseen variations of the rotation nuisance factor.

Table~\ref{tab:mnist_rot} summarizes the results, showing that our adversarial model, which is trained without any $z$ information, not only performs better than the baseline ablation versions but also outperforms state-of-the art methods, which use supervised information about the rotation angle. The difference in $A_y$ is especially notable for the cases where $\theta \not \in \Theta$. Results on $A_z$ show that our model discards more information about $\theta$ than previous works even though prior art uses $\theta$ information during training. The information about $\theta$ migrates to $e_2$, indicated by the accuracy of predicting it from $e_2$ being $0.77$. Figure~\ref{fig:recon_mnist_rot} shows results of reconstructing $x$ from $e_1$ and $e_2$ generated in the same way as Extended Yale-B above. The figures show that reconstructions from $e_1$ reflect the digit class but contain no information about $\theta$, while those from $e_2$ exhibit the inverse. Figure~\ref{fig:tsne_mnist_rot} shows t-SNE visualization of raw MNIST-ROT images and $e_1$ learned by our model. While raw data tends to cluster by $\theta$, $e_1$ shows near-perfect grouping based on the digit-class. We further visualize the $e_1$ embedding learned by the proposed model and the baseline $B_0$, which models the classifier $x \dashedrightarrow h \dashedrightarrow y$, to investigate the effectiveness of invariance induction by exclusion versus inclusion, respectively. Both the models were trained on digits rotated by $\theta \in \Theta$ and  t-SNE visualizations were generated for $\theta \in \{\pm55\}$. Figure~\ref{fig:e1_55} shows the results. As evident, $e_1$ learned by the proposed model shows no clustering by the rotation angle, while that learned by $B_0$ does, with encodings of some digit classes forming multiple clusters corresponding to rotation angles.

{
\def \fs {0.4}
\def \sfs {0.725}
\begin{figure*}
\centering
\captionsetup[subfigure]{aboveskip=2pt,belowskip=2pt}
\begin{subfigure}{\fs\textwidth}
\centering
\includegraphics[width=\sfs\textwidth]{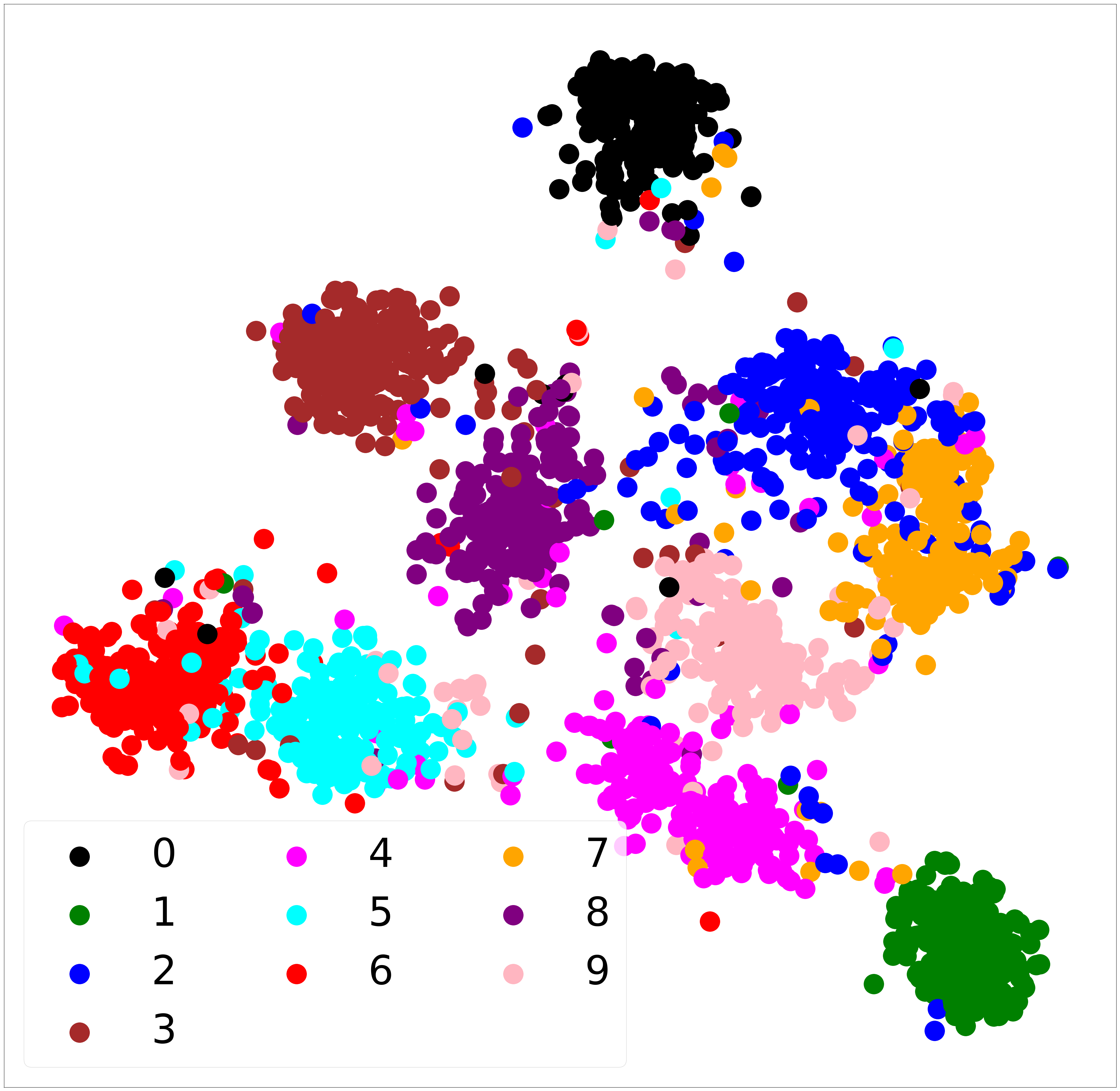}
\caption{}
\end{subfigure}
\begin{subfigure}{\fs\textwidth}
\centering
\includegraphics[width=\sfs\textwidth]{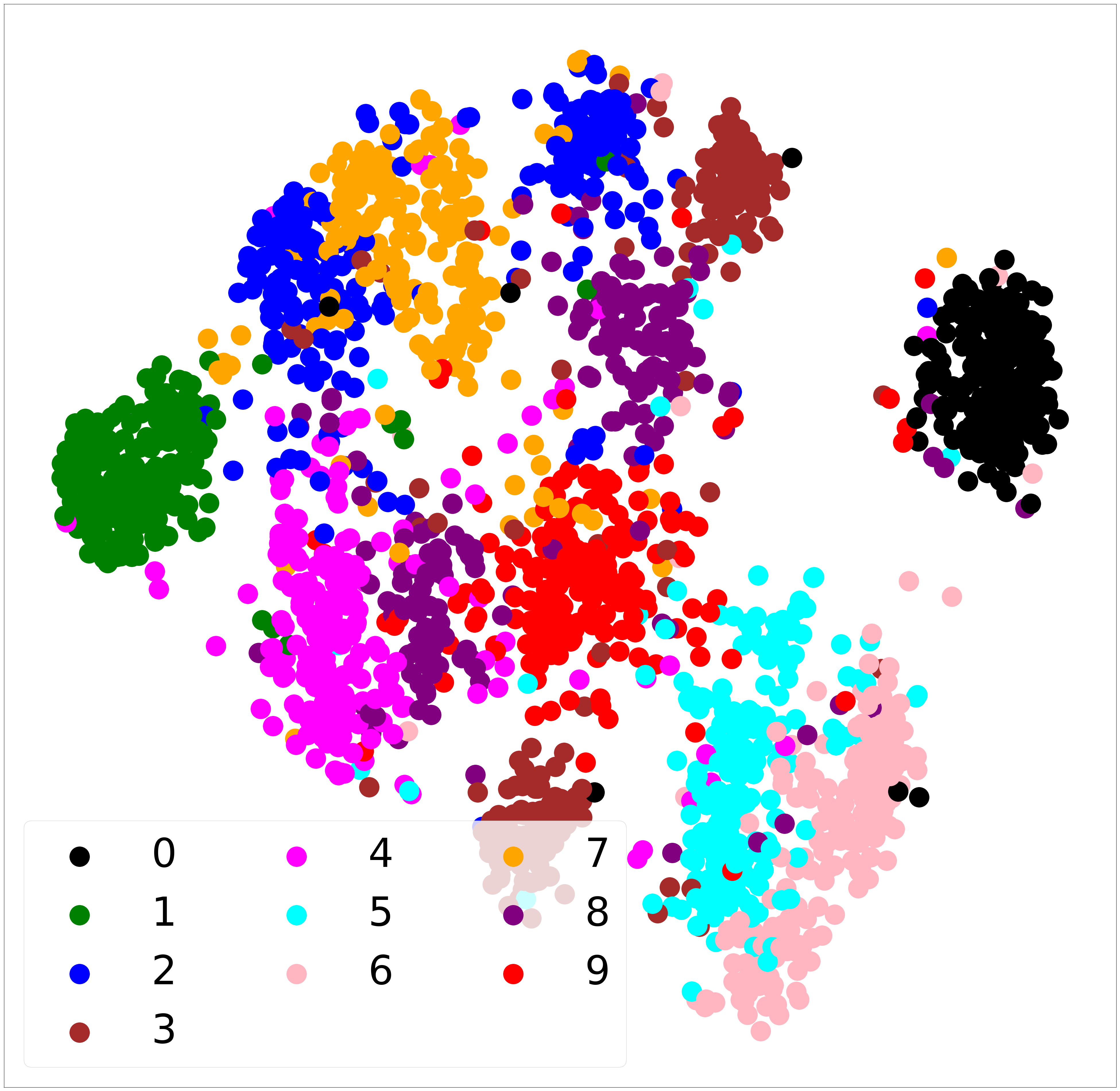}
\caption{}
\end{subfigure}
\begin{subfigure}{\fs\textwidth}
\centering
\includegraphics[width=\sfs\textwidth]{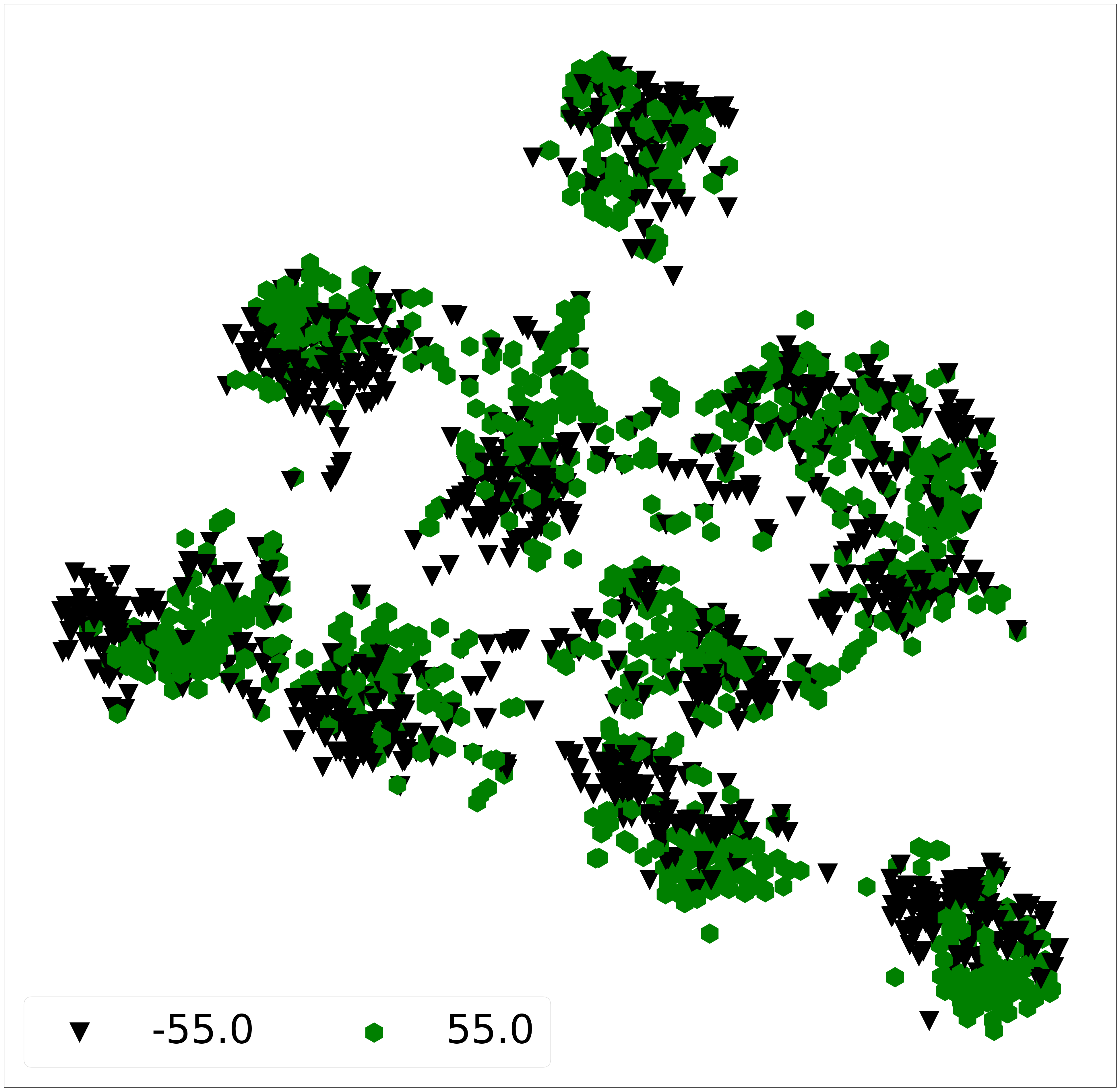}
\caption{}
\end{subfigure}
\begin{subfigure}{\fs\textwidth}
\centering
\includegraphics[width=\sfs\textwidth]{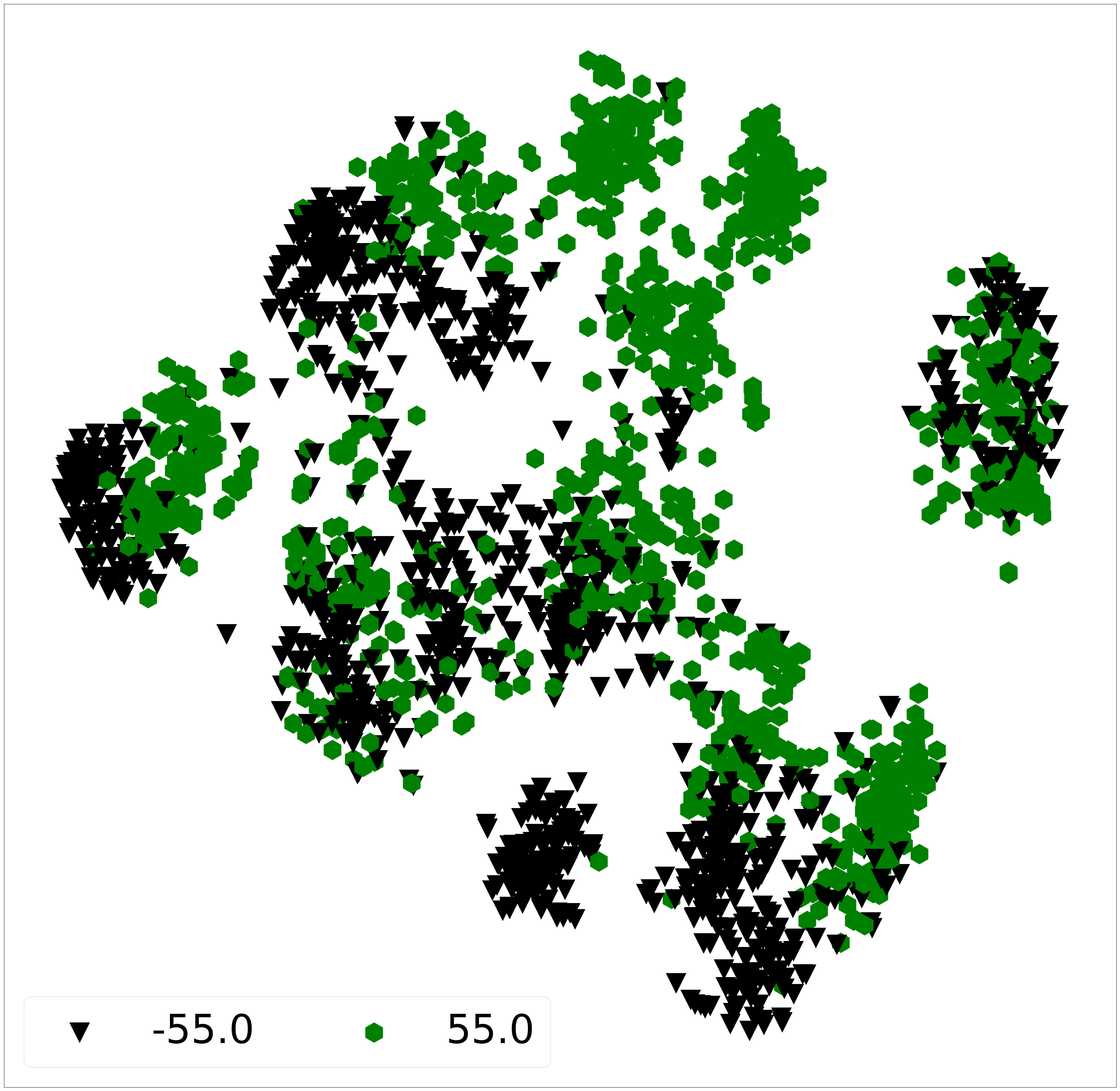}
\caption{}
\end{subfigure}
\vspace{-8pt}
\caption{\label{fig:e1_55}t-SNE visualization of MNIST-ROT $e_1$ embedding for UnifAI (a) \& (c), and baseline model $B_0$ (b) \& (d). Models were trained on $\theta \in \{0, \pm22.5, \pm45\}$. Visualization is presented for $\theta = \pm55$. $B_0$ embeddings show sub-clusters of $\theta$ within each digit cluster, such that $\theta$ information is easily separable (d). The UnifAI embedding $e_1$ does not show any grouping by $\theta$.}
\end{figure*}
}

\setlength{\tabcolsep}{1.55em} 
\begin{table*}
\makegapedcells
\centering
\caption{\label{tab:mnist_dil}MNIST-DIL -- Accuracy of predicting $y$ ($A_y$). $\kappa = -2$ \ represents erosion with kernel-size of $2$.}
\vspace{-7pt}
\begin{tabular}{ c ?{1.5pt} c ?{0.75pt} c ?{0.75pt} c ?{0.75pt} c ?{1.5pt} c ?{0.25pt} c ?{0.25pt} c }
  \hbline
  $\boldsymbol{\kappa}$ & \textbf{NN+MMD~\cite{bib:nnmmd}} & \textbf{VFAE~\cite{bib:vfae}} & \textbf{CAI~\cite{bib:sai}} & \textbf{CVIB~\cite{bib:cvib}} & \textbf{UnifAI (ours)} & $\boldsymbol{B_1}$ & $\boldsymbol{B_0}$ \\
  \hbline
  -2 & 0.870 & 0.807 & 0.816 & 0.844 & \textbf{0.880} & 0.870 & 0.872 \\
  2 & 0.944 & 0.916 & 0.933 & 0.933 & \textbf{0.958} & 0.940 & 0.942 \\
  3 & 0.855 & 0.818 & 0.795 & 0.846 & \textbf{0.874} & 0.853 & 0.847 \\
  4 & 0.574 & 0.548 & 0.519 & 0.586 & \textbf{0.606} & 0.550 & 0.534 \\
  \hbline
\end{tabular}
\vspace{-10pt}
\end{table*}

\begin{figure}
\centering
\includegraphics[width=0.415\textwidth]{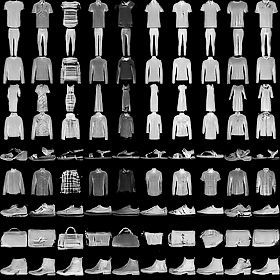}
\vspace{-5pt}
\caption{\label{fig:fashion_mnist_gen}Random samples generated using the BiCoGAN trained on Fashion-MNIST. Rows indicate classes. The latent embedding (style) is fixed for each column.}
\vspace{-25pt}
\end{figure}





\paragraph*{MNIST-DIL} We create this variant of MNIST by eroding or dilating MNIST digits using various kernel-sizes ($\kappa$). We use models trained on MNIST-ROT to report evaluation results on this dataset, to show the advantage of unsupervised invariance induction in cases where certain $z$ are not annotated in the training data. Thus, information about these $z$ cannot be used to train supervised invariance models.

Table~\ref{tab:mnist_dil} summarizes the results of this experiment. The results show significantly better performance of our model compared to all the baselines. More notably, prior works perform \emph{significantly} worse than our baseline models, indicating that the supervised approach of invariance induction can worsen performance with respect to nuisance factors not accounted for during training.

\subsection{Learning invariance to arbitrary nuisance factors by leveraging Generative Adversarial Networks}

Algorithmic generation of synthetic data for augmentation allows the generation of specific forms of variation of data that they are designed for. However, it is very difficult to account for all possible variations of data using such approaches. In light of this, Generative Adversarial Networks (GANs)~\cite{bib:gan} have recently been employed for data augmentation and have provided significant gains on the final prediction performance of the supervised task~\cite{bib:gan_aug}. The ability of generating massive amounts of arbitrary variations of data using GANs combined with the proposed invariance induction framework provides a novel approach for the development of robust features that are, in theory, invariant to all forms of nuisance factors in data (with respect to the supervised task) that define the underlying generative model parameterized by the GAN. We evaluate this experimental setting on two datasets -- Fashion-MNIST~\cite{bib:fashion_mnist} and Omniglot~\cite{bib:omniglot}. We report results of three configurations -- (1) $B_0$ (baseline model composed of $Enc$ and $Pred$) trained on real training data, (2) $B_0$ trained on real and generated data (augmented dataset), and (3) the proposed model trained on the augmented dataset. Since the data is generated with arbitrary variations of the latent nuisance factors, $z$ is not easily quantifiable for this experiment. We present results of these configurations on real testing data as well as \textit{extreme samples}, which are difficult examples sampled far from modes of the latent distribution of the GAN models.

{
\def \fs {0.32}
\def \sfs {0.9}
\begin{figure*}
\centering
\captionsetup[subfigure]{font=bf}
\begin{subfigure}{\fs\textwidth}
\centering
\caption*{$\boldsymbol{B_0}$ trained on real data}
\includegraphics[width=\sfs\textwidth]{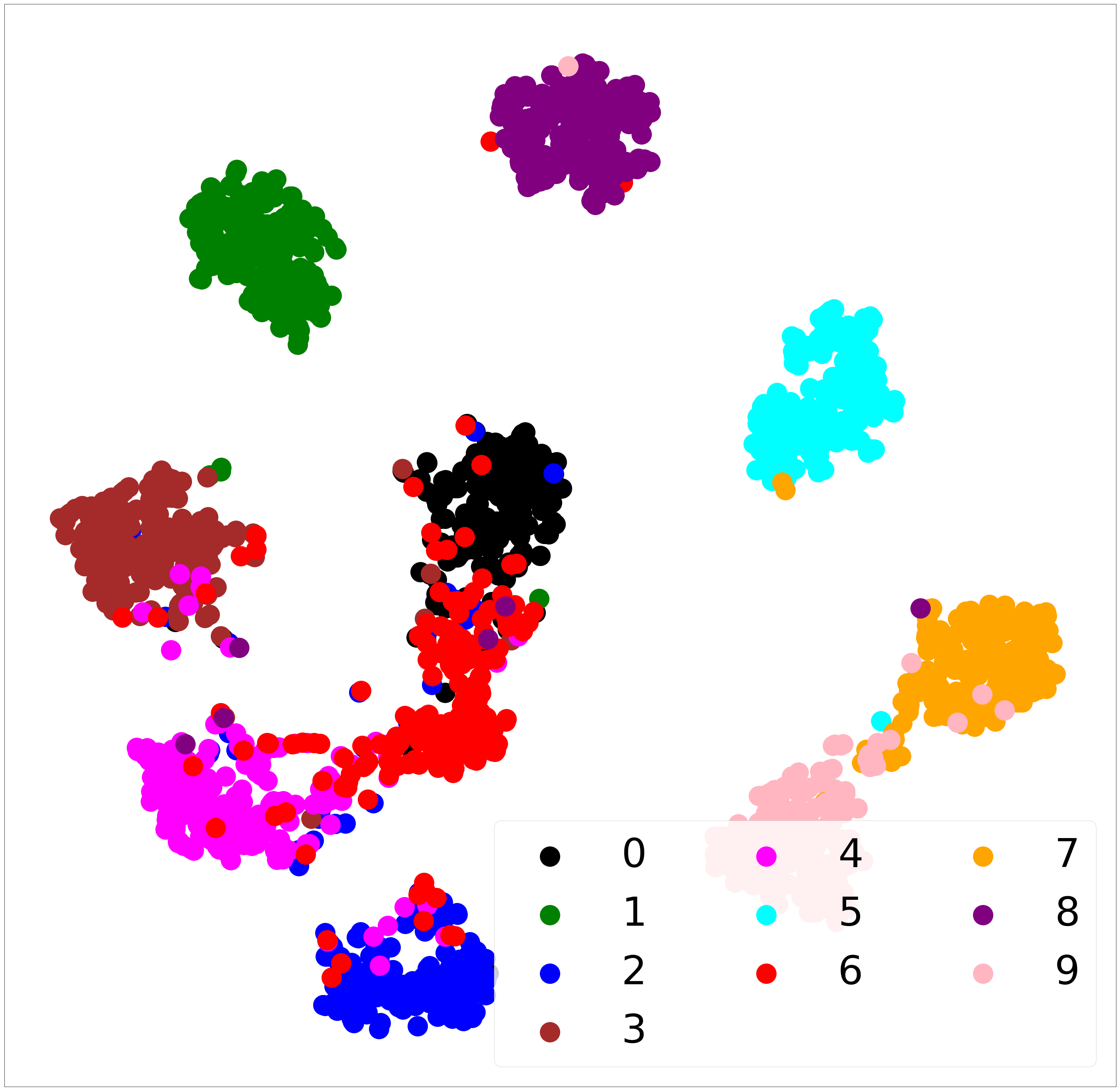}
\end{subfigure}\hfill%
\begin{subfigure}{\fs\textwidth}
\centering
\caption*{$\boldsymbol{B_0}$ trained on aug. data}
\includegraphics[width=\sfs\textwidth]{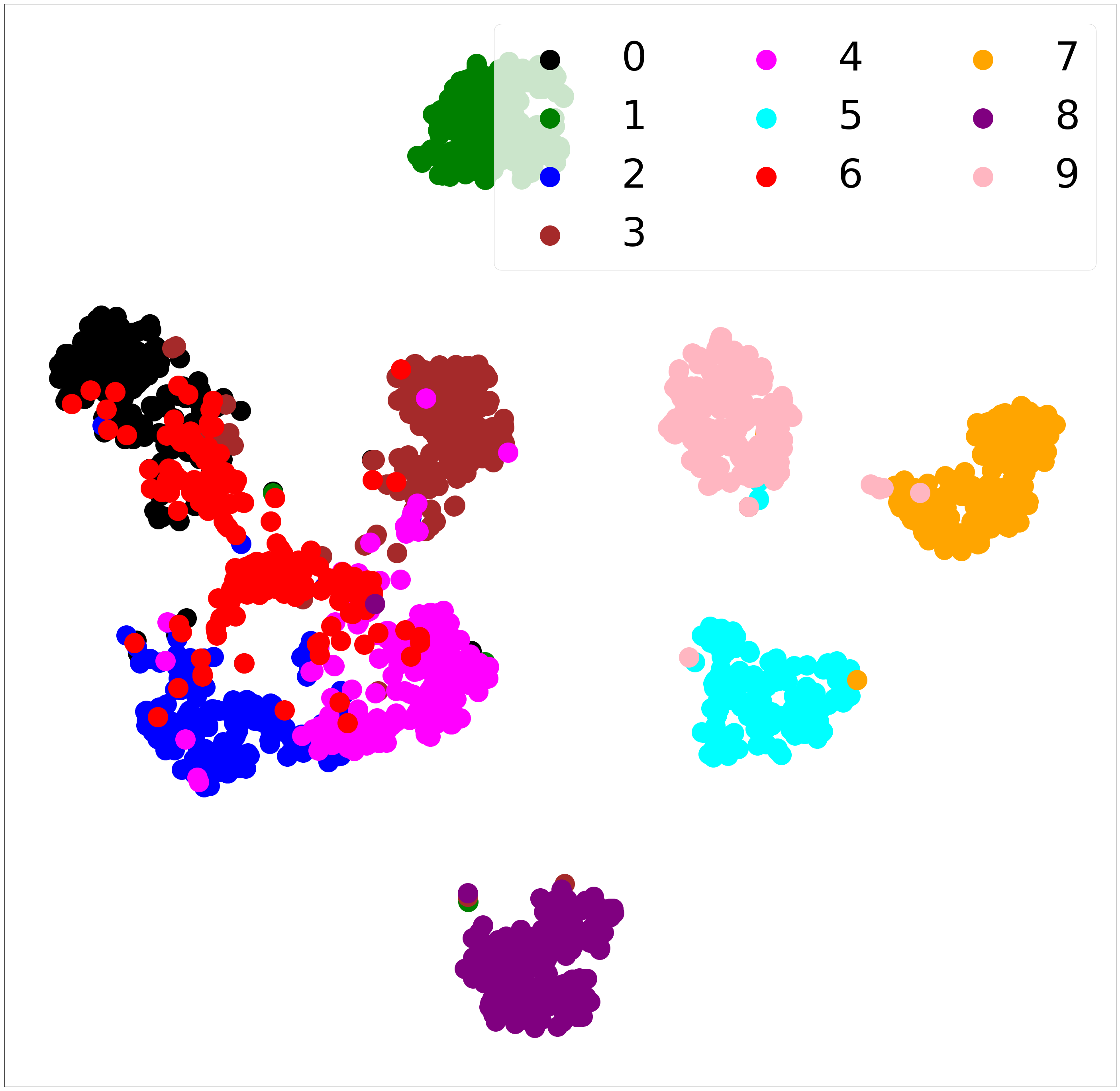}
\end{subfigure}\hfill%
\begin{subfigure}{\fs\textwidth}
\centering
\caption*{UnifAI trained on aug. data}
\includegraphics[width=\sfs\textwidth]{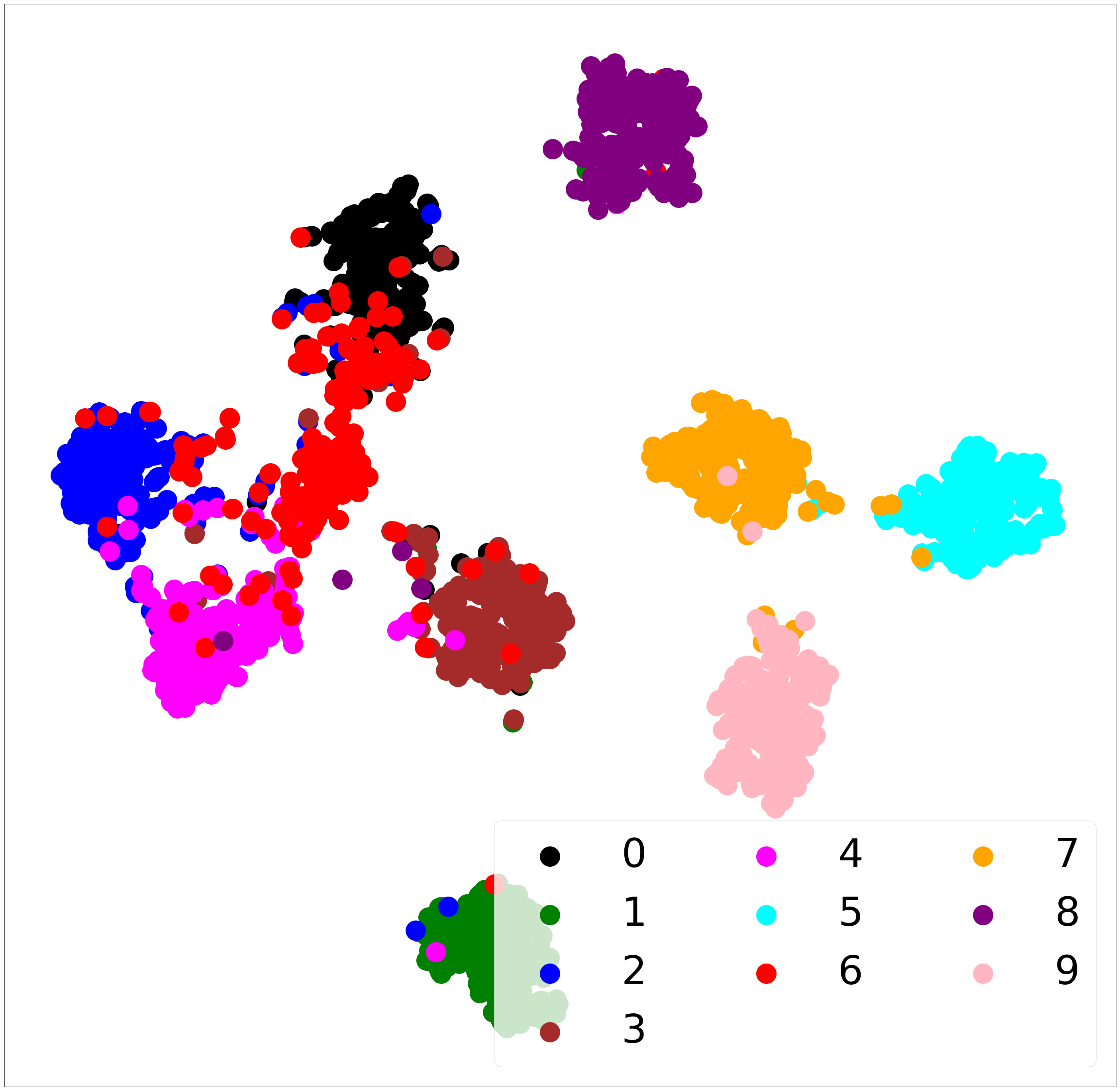}
\end{subfigure}
\begin{subfigure}{\fs\textwidth}
\centering
\caption*{}
\includegraphics[width=\sfs\textwidth]{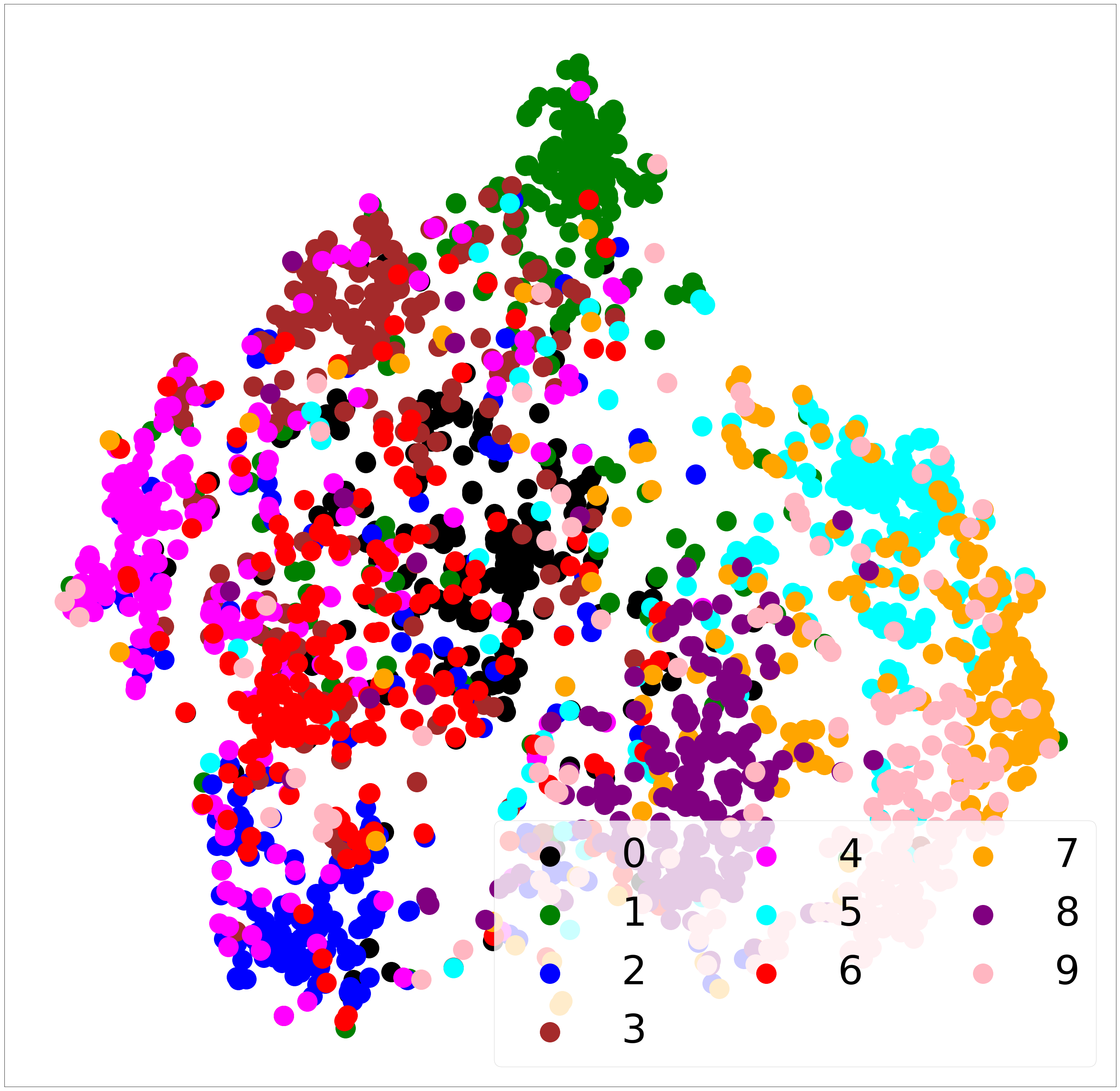}
\end{subfigure}\hfill%
\begin{subfigure}{\fs\textwidth}
\centering
\caption*{}
\includegraphics[width=\sfs\textwidth]{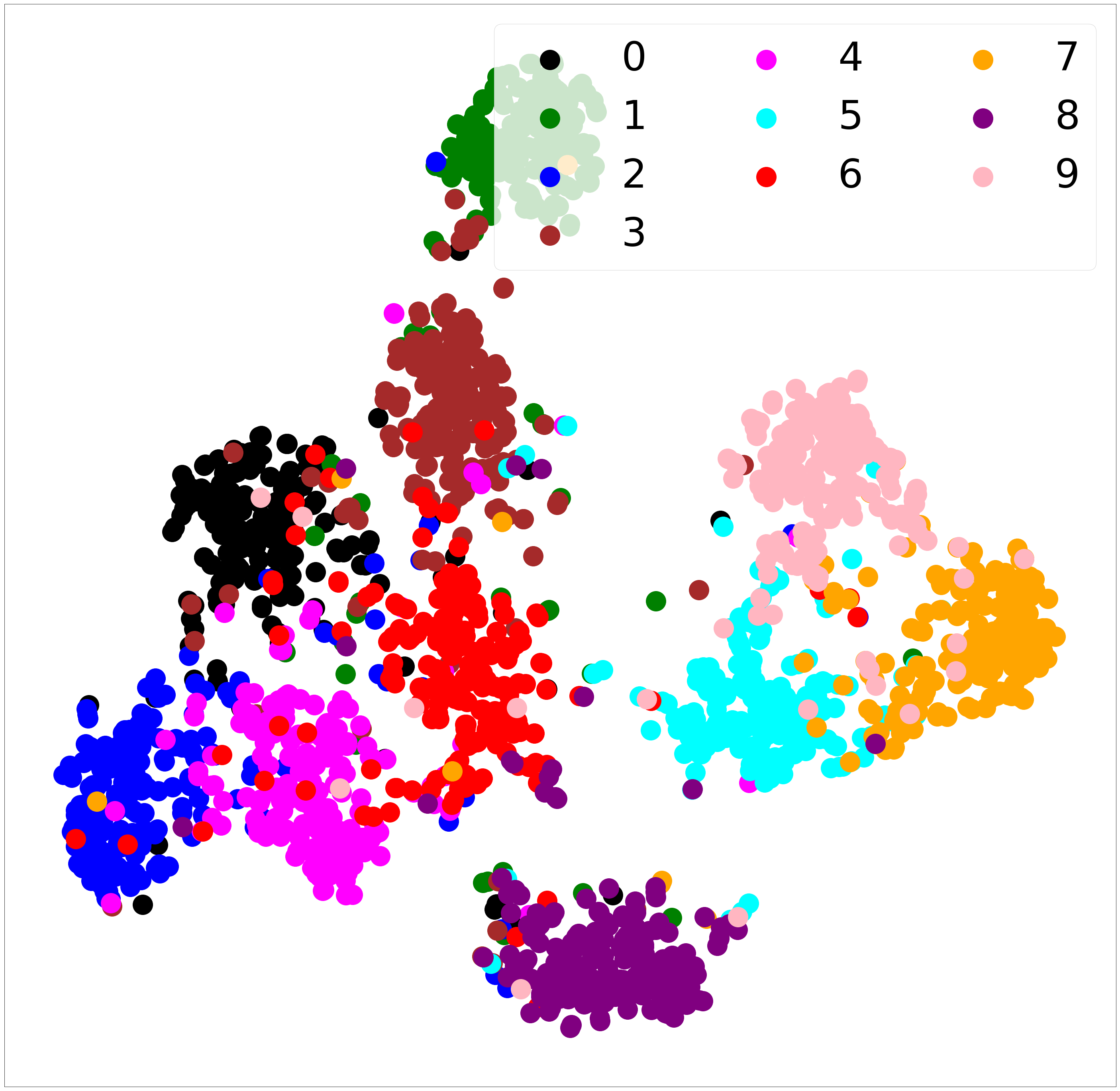}
\end{subfigure}\hfill%
\begin{subfigure}{\fs\textwidth}
\centering
\caption*{}
\includegraphics[width=\sfs\textwidth]{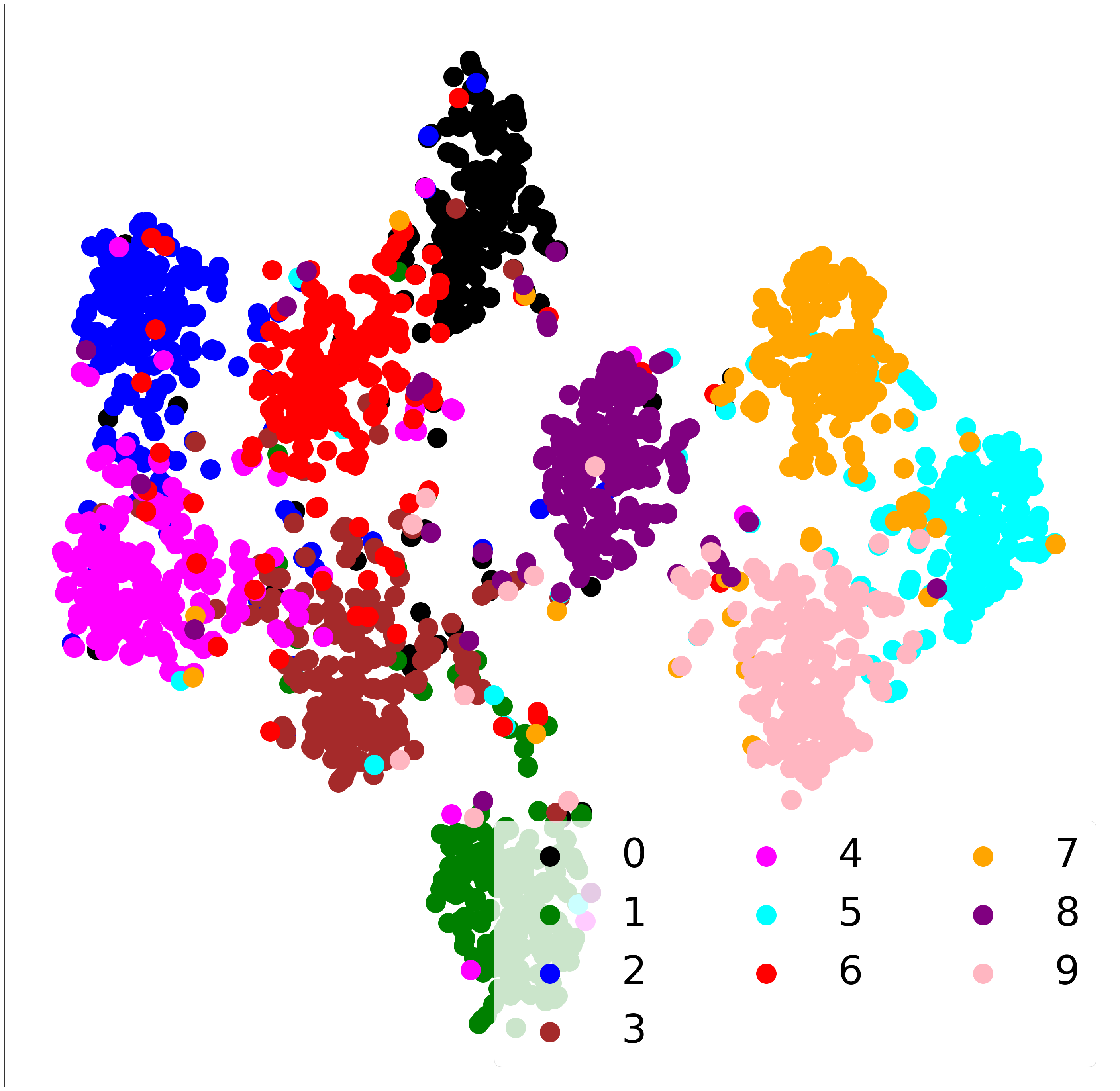}
\end{subfigure}
\caption{\label{fig:tsne_fashion_mnist}t-SNE visualization of Fashion-MNIST $e_1$ embedding. The first row shows the visualization for real test data and the second shows that for extreme test samples. The embedding of the extreme test samples from the $B_0$ model trained on real data is scattered with vague clustering by the clothing-class. Training $B_0$ with the augmented dataset makes the clustering of extreme samples cleaner. This clustering improves further in the case of the UnifAI model trained with augmented data.}
\vspace{-8pt}
\end{figure*}
}

\setlength{\tabcolsep}{0.48em} 
\begin{table}
\makegapedcells
\centering
\caption{\label{tab:fashion_mnist}Fashion-MNIST -- Accuracy of predicting $y$ ($A_y$)}
\begin{tabular}{ c ?{1.5pt} c ?{0.75pt} c ?{1.5pt} c }
  \hbline
  \textbf{Test-set} & $\boldsymbol{B_0}$\textbf{ + Real} & $\boldsymbol{B_0}$\textbf{ + Augmented} & \textbf{UnifAI + Augmented} \\
  \hbline
  Real & 0.918 & 0.922 & \textbf{0.934} \\
  Extreme & 0.640 & 0.876 & \textbf{0.889} \\
  \hbline
\end{tabular}
\vspace{-4pt}
\end{table}

\begin{figure}
\centering
\includegraphics[width=0.4\textwidth]{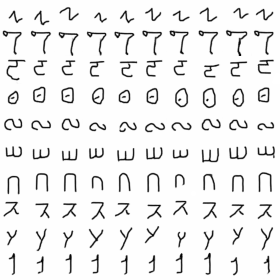}
\vspace{-5pt}
\caption{\label{fig:omniglot_gen}Random samples generated using DAGAN trained on Omniglot. Rows indicate $10$ randomly sampled classes. The latent embedding (style) is fixed for each column.}
\vspace{-10pt}
\end{figure}

\paragraph*{Fashion-MNIST} This dataset contains grayscale images of 10 kinds of clothing. It was designed as a more challenging replacement of the MNIST dataset for benchmarking machine learning models. The target $y$ in the supervised task is the type of clothing (e.g. trouser, coat, etc.), whereas nuisance factors include all elements of style that are not particular to (discriminative of) specific clothing classes. We trained a Bidirectional Conditional GAN (BiCoGAN)~\cite{bib:bicogan} on the training set. Figure~\ref{fig:fashion_mnist_gen} qualitatively shows the performance of the BiCoGAN through randomly sampled images. We sampled the generated instances for training the $B_0$ and UnifAI models two standard deviations away from the mean of the latent distribution. This was done to avoid generating samples that are very similar to real examples and thus have very little variation with respect to the real training dataset. Extreme examples for testing were sampled three standard deviations away from the distribution mean. We model $Enc$ as a neural network composed of two convolution layers followed by two fully-connected layers, $Pred$ as two fully-connected layers, $Dec$ as three convolution layers and the disentanglers as two fully-connected layers.

Table~\ref{tab:fashion_mnist} summarizes the results of our experiments. As evident, training $B_0$ with augmented data generated using the GAN model improves the prediction accuracy on the real test data as well as extreme examples, as compared to training $B_0$ with only real training data. However, the configuration with the proposed invariance induction framework trained with the augmented dataset achieves the best performance. Figure~\ref{fig:tsne_fashion_mnist} shows the t-SNE visualization of the embedding used for classification for real and extreme test samples. The figure shows that the embedding of real data does not change much across the three configurations. However, that of extreme samples improves progressively in the order: $B_0$ trained on real data, $B_0$ trained on augmented data, and the proposed framework trained on augmented data. This correlates with the quantitative results in Table~\ref{tab:fashion_mnist}.




\setlength{\tabcolsep}{0.48em} 
\begin{table}
\makegapedcells
\centering
\caption{\label{tab:omniglot}Omniglot -- Accuracy of predicting $y$ ($A_y$)}
\begin{tabular}{ c ?{1.5pt} c ?{0.75pt} c ?{1.5pt} c }
  \hbline
  \textbf{Test-set} & $\boldsymbol{B_0}$\textbf{ + Real} & $\boldsymbol{B_0}$\textbf{ + Augmented} & \textbf{UnifAI + Augmented} \\
  \hbline
  Real & 0.674 & 0.725 & \textbf{0.740} \\
  Extreme & 0.414 & 0.535 & \textbf{0.558} \\
  \hbline
\end{tabular}
\end{table}

\paragraph*{Omniglot} This is a dataset of 1,623 different handwritten characters from 50 different alphabets. The target $y$ is the character-type whereas elements of handwriting style that are not discriminative of $y$ are considered as nuisance factors. We trained the Data Augmentation GAN (DAGAN)~\cite{bib:gan_aug} using the official code\footnote{\href{https://www.github.com/AntreasAntoniou/DAGAN}{https://www.github.com/AntreasAntoniou/DAGAN}} available for this dataset. Figure~\ref{fig:omniglot_gen} qualitatively shows the performance of the DAGAN through randomly sampled images. The generated dataset for training the models was sampled one standard deviation away from the mean of the latent distribution, whereas extreme examples for testing were sampled two standard deviations away from the mean. We used a neural network composed of three convolution layers followed by two fully-connected layers for $Enc$, two fully-connected layers for $Pred$, three convolution layers for $Dec$, and two fully-connected layers for the disentanglers.

Table~\ref{tab:omniglot} shows the results of our experiments. As with the case of Fashion-MNIST above, training $B_0$ with the augmented dataset leads to better classification accuracy on not only the real test dataset but also the extreme samples, as compared to $B_0$ trained with only the real training dataset. The proposed invariance framework trained with the augmented dataset, however, achieves the best performance, further supporting the effectiveness of the proposed framework in leveraging GANs for learning invariance to arbitrary nuisance factors.

\subsection{Domain Adaptation}

Domain adaptation has been treated as an invariance induction task recently (\citem{bib:dann,bib:vfae}) where the goal is to make the prediction task invariant to the ``domain'' information. We evaluate the performance of our model at domain adaptation on the Amazon Reviews dataset~\cite{bib:amzrev} using the same preprocessing as~\cite{bib:vfae}. The dataset contains text reviews on products in four domains -- ``books'', ``dvd'', ``electronics'', and ``kitchen''. Each review is represented as a feature vector of unigram and bigram counts. The target $y$ is the sentiment of the review -- either positive or negative. We use the same experimental setup as (\citem{bib:dann,bib:vfae}) where the model is trained on one domain and tested on another, thus creating $12$ source-target combinations. We design the architectures of the encoder and the decoder in our model to be similar to those of VFAE, as presented in~\cite{bib:vfae}. Table~\ref{tab:amzrev} shows the results of our model trained without $D_z$ and supervised state-of-the-art methods VFAE and Domain Adversarial Neural Network (DANN)~\cite{bib:dann}, which use $z$ labels during training. The results of the prior works are quoted directly from~\cite{bib:vfae}. The results show that our model outperforms both VFAE and DANN at nine out of the twelve tasks. Thus, our model can also be used effectively for domain adaptation.

\subsection{Fair Representation learning}

Learning invariance to biasing factors requires information about these factors to discard from the prediction process because they are correlated with the prediction target and cannot be removed in an unsupervised way. Hence, we use the full unified framework, which includes the $z$-discriminator $D_z$, for this task. We provide results of our model and prior state-of-the-art methods (NN+MMD, VFAE, CAI, and CVIB) on Adult (\citem{bib:adult,bib:uci}) and German (\citem{bib:adult,bib:uci}) datasets, which are used popularly in evaluating fair representation frameworks (\citem{bib:vfae,bib:sai,bib:cvib}). We used the same preprocessed versions of these datasets as~\cite{bib:vfae}.

\setlength{\tabcolsep}{0.6em} 
\begin{table}
\makegapedcells
\centering
\caption{\label{tab:amzrev}Amazon Reviews dataset -- Accuracy of predicting $y$ from $e_1$ ($A_y$). CAI~\cite{bib:sai} and DANN~\cite{bib:dann} are the same model.}
\begin{tabular}{ c ?{1.5pt} c ?{0.75pt} c ?{1.5pt} c }
  \hbline
  \textbf{Source - Target} & \textbf{DANN~\cite{bib:dann}} & \textbf{VFAE~\cite{bib:vfae}} & \textbf{UnifAI (Ours)} \\
  \hbline
  books - dvd & 0.784 & 0.799 & \textbf{0.820} \\
  books - electronics & 0.733 & \textbf{0.792} & 0.764 \\
  books - kitchen & 0.779 & \textbf{0.816} & 0.791 \\
  \hline
  dvd - books & 0.723 & 0.755 & \textbf{0.798} \\
  dvd - electronics & 0.754 & 0.786 & \textbf{0.790} \\
  dvd - kitchen & 0.783 & 0.822 & \textbf{0.826} \\
  \hline
  electronics - books & 0.713 & 0.727 & \textbf{0.734} \\
  electronics - dvd & 0.738 & \textbf{0.765} & 0.740 \\
  electronics - kitchen & 0.854 & 0.850 & \textbf{0.890} \\
  \hline
  kitchen - books & 0.709 & 0.720 & \textbf{0.724} \\
  kitchen - dvd & 0.740 & 0.733 & \textbf{0.745} \\
  kitchen - electronics & 0.843 & 0.838 & \textbf{0.859} \\
  \hbline
\end{tabular}
\end{table}

{
\def \fs {0.32}
\def \sfs {0.9}
\begin{figure*}[h]
\centering
\begin{subfigure}{\fs\textwidth}
\centering
\includegraphics[width=\sfs\textwidth]{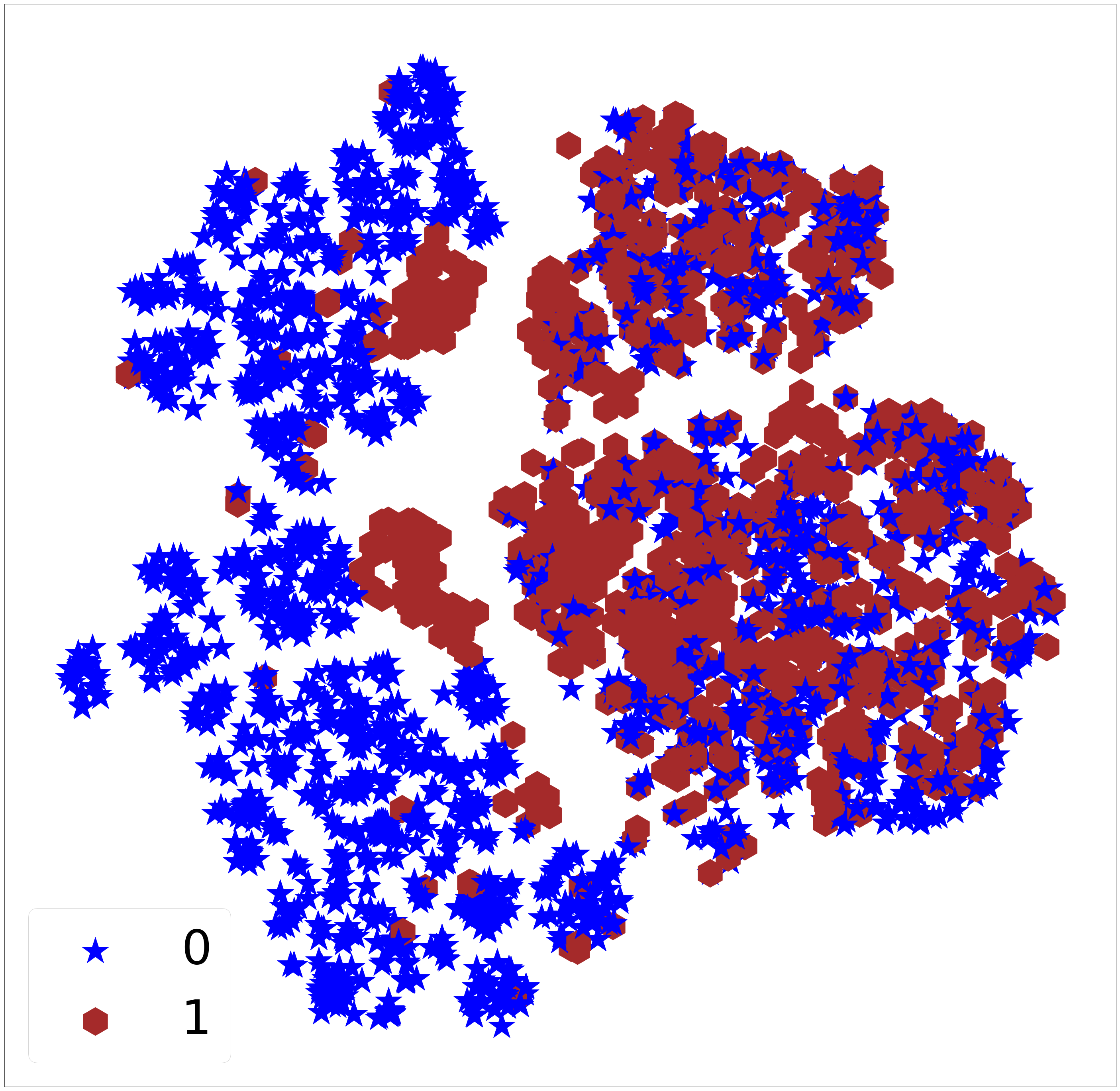}
\caption{Raw data}
\end{subfigure}
\begin{subfigure}{\fs\textwidth}
\centering
\includegraphics[width=\sfs\textwidth]{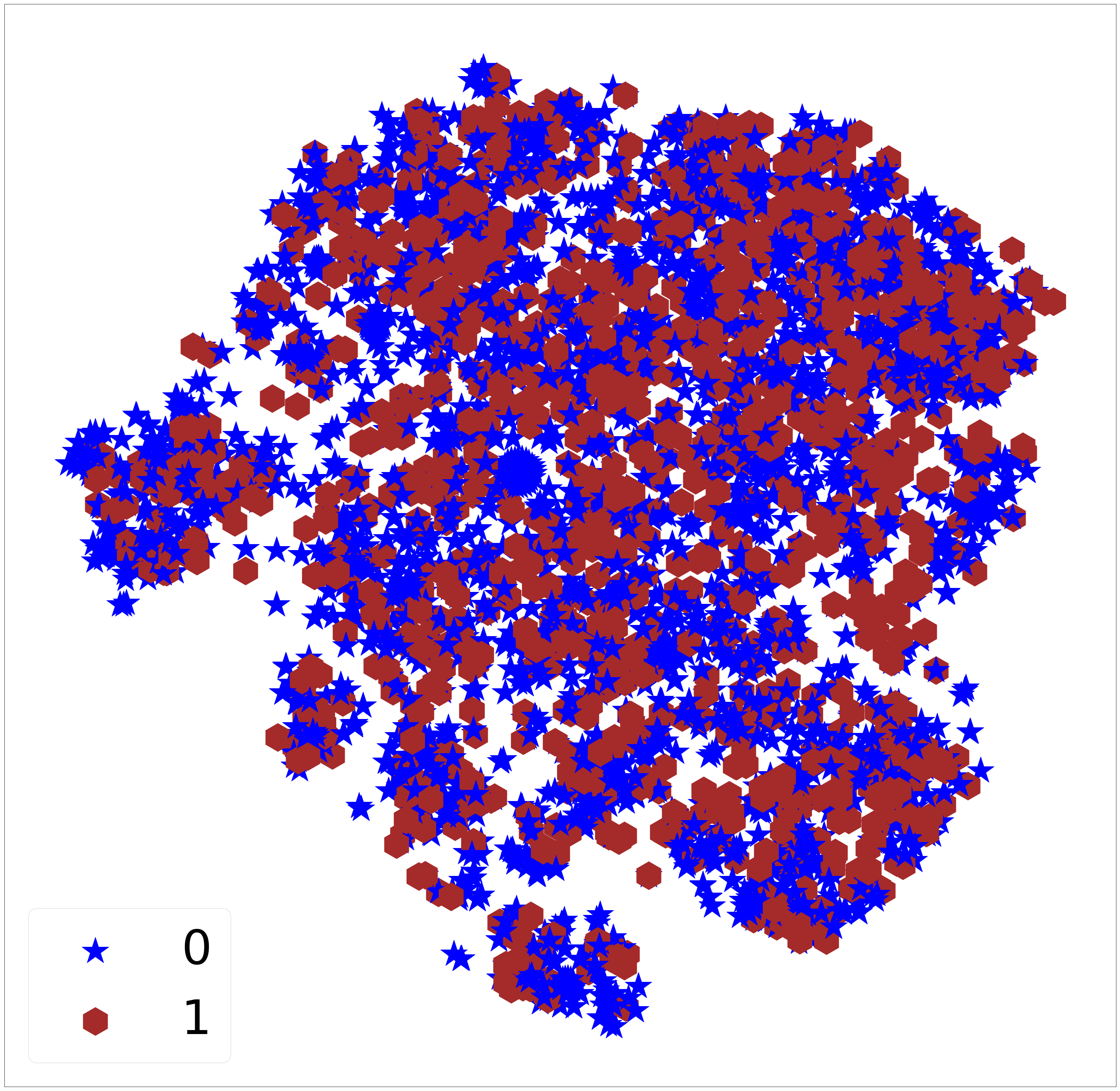}
\caption{$e_1$ embedding}
\end{subfigure}
\begin{subfigure}{\fs\textwidth}
\centering
\includegraphics[width=\sfs\textwidth]{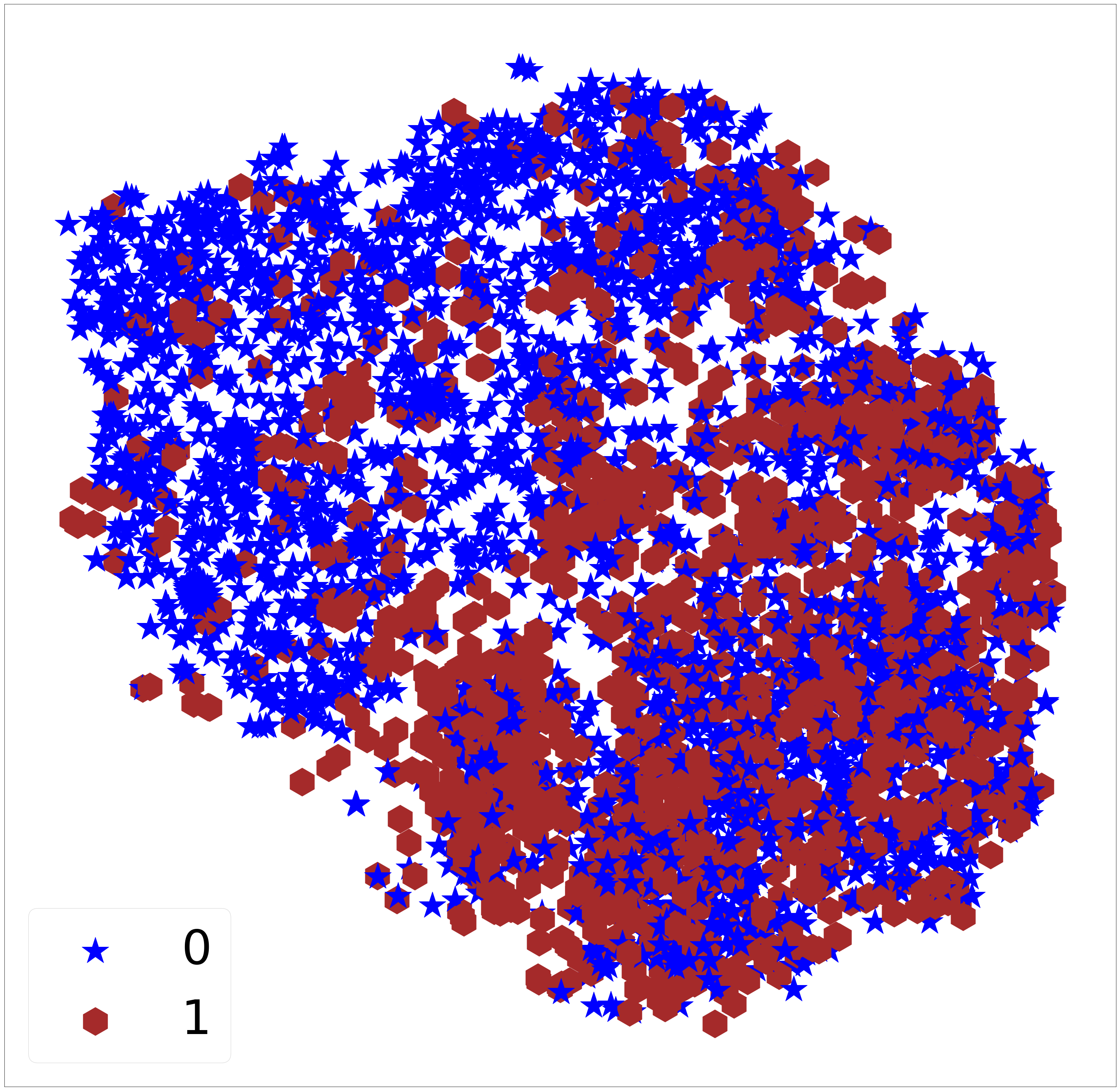}
\caption{$e_2$ embedding}
\end{subfigure}
\caption{\label{fig:tsne_adult}Adult dataset -- t-SNE visualization. Labels indicate the biasing factor -- age. Raw data clusters by age, showing the bias. $e_1$ does not cluster by age, as desired for fairness, while $e_2$ does, showing the migration of bias to $e_2$.}
\end{figure*}

\setlength{\tabcolsep}{1.75em} 
\begin{table*}
\makegapedcells
\centering
\caption{\label{tab:adult}Results on Adult dataset}
\begin{tabular}{ c ?{1.5pt} c ?{0.75pt} c ?{0.75pt} c ?{0.75pt} c ?{1.5pt} c }
  \hbline
  \textbf{Metric} & \textbf{NN+MMD~\cite{bib:nnmmd}} & \textbf{VFAE~\cite{bib:vfae}} & \textbf{CAI~\cite{bib:sai}} & \textbf{CVIB~\cite{bib:cvib}} & \textbf{UnifAI (ours)} \\
  \hbline
  $A_y$ & 0.75 & 0.76 & 0.83 & 0.69 & \textbf{0.84} \\
  $A_z$ & \textbf{0.67} & \textbf{0.67} & 0.89 & 0.68  & \textbf{0.67} \\
  \hbline
\end{tabular}
\end{table*}
\setlength{\tabcolsep}{\defaulttabcolsep} 

\begin{figure*}
\begin{subfigure}{\fs\textwidth}
\centering
\includegraphics[width=\sfs\textwidth]{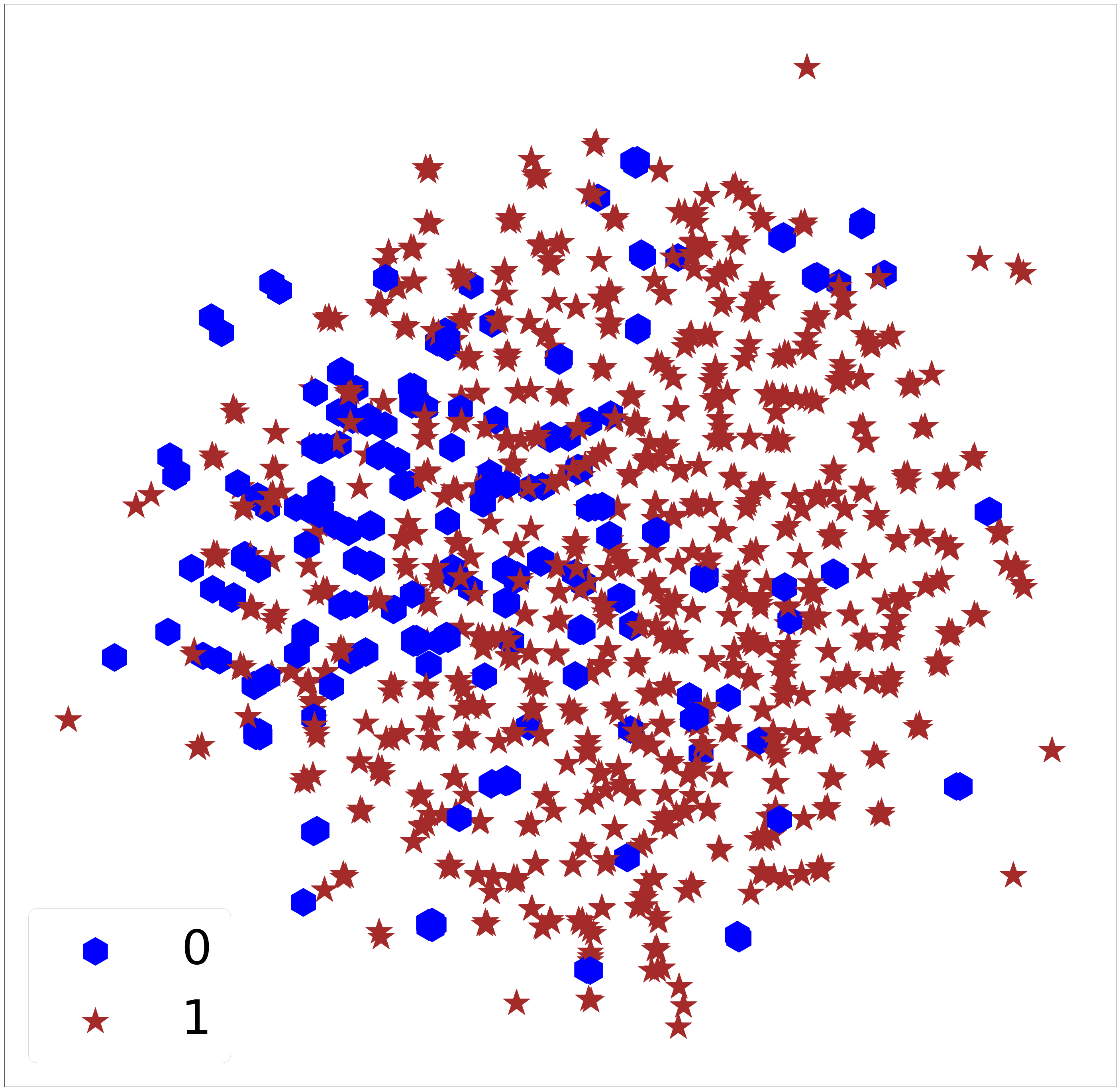}
\caption{Raw data}
\end{subfigure}
\begin{subfigure}{\fs\textwidth}
\centering
\includegraphics[width=\sfs\textwidth]{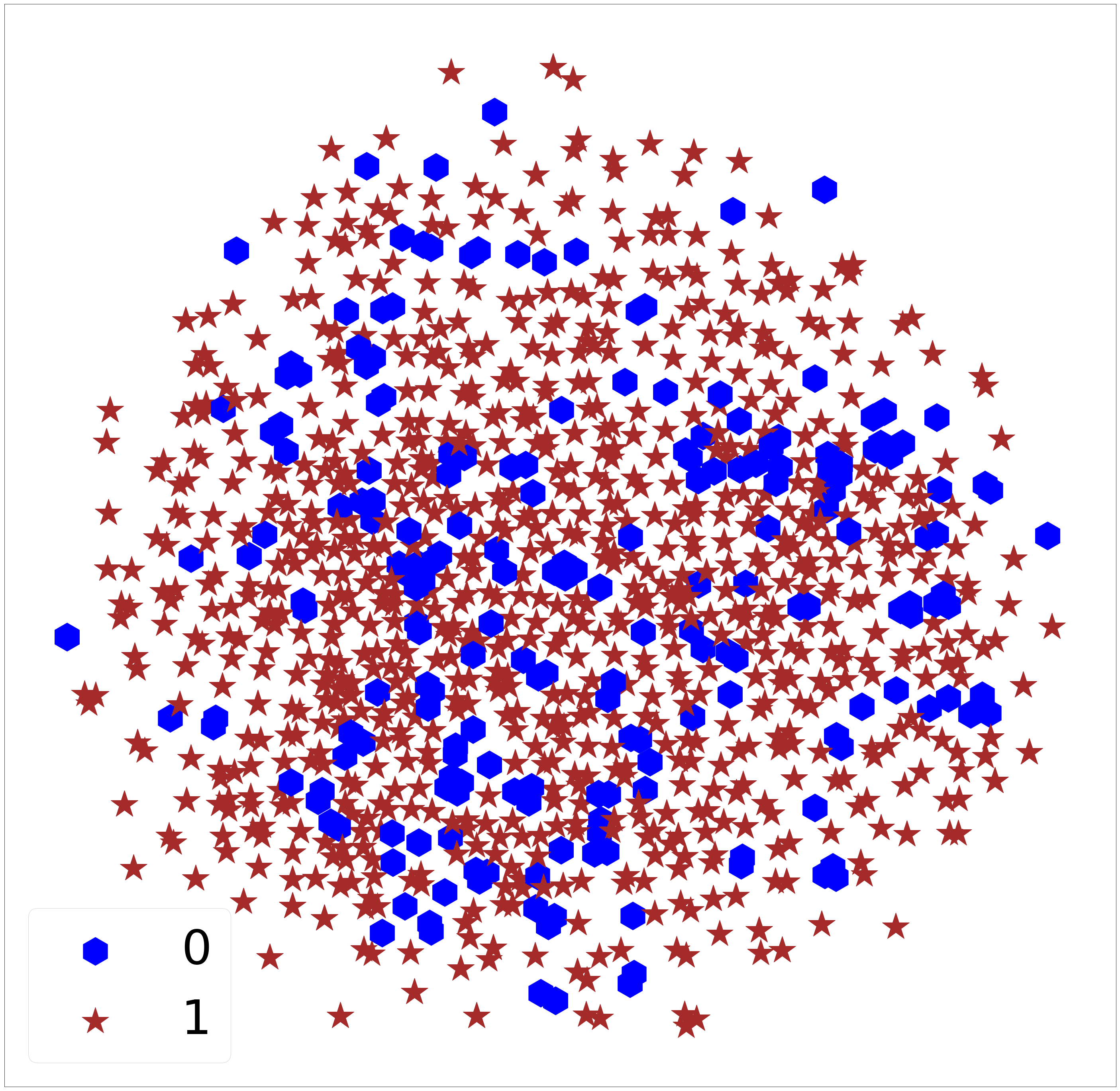}
\caption{$e_1$ embedding}
\end{subfigure}
\begin{subfigure}{\fs\textwidth}
\centering
\includegraphics[width=\sfs\textwidth]{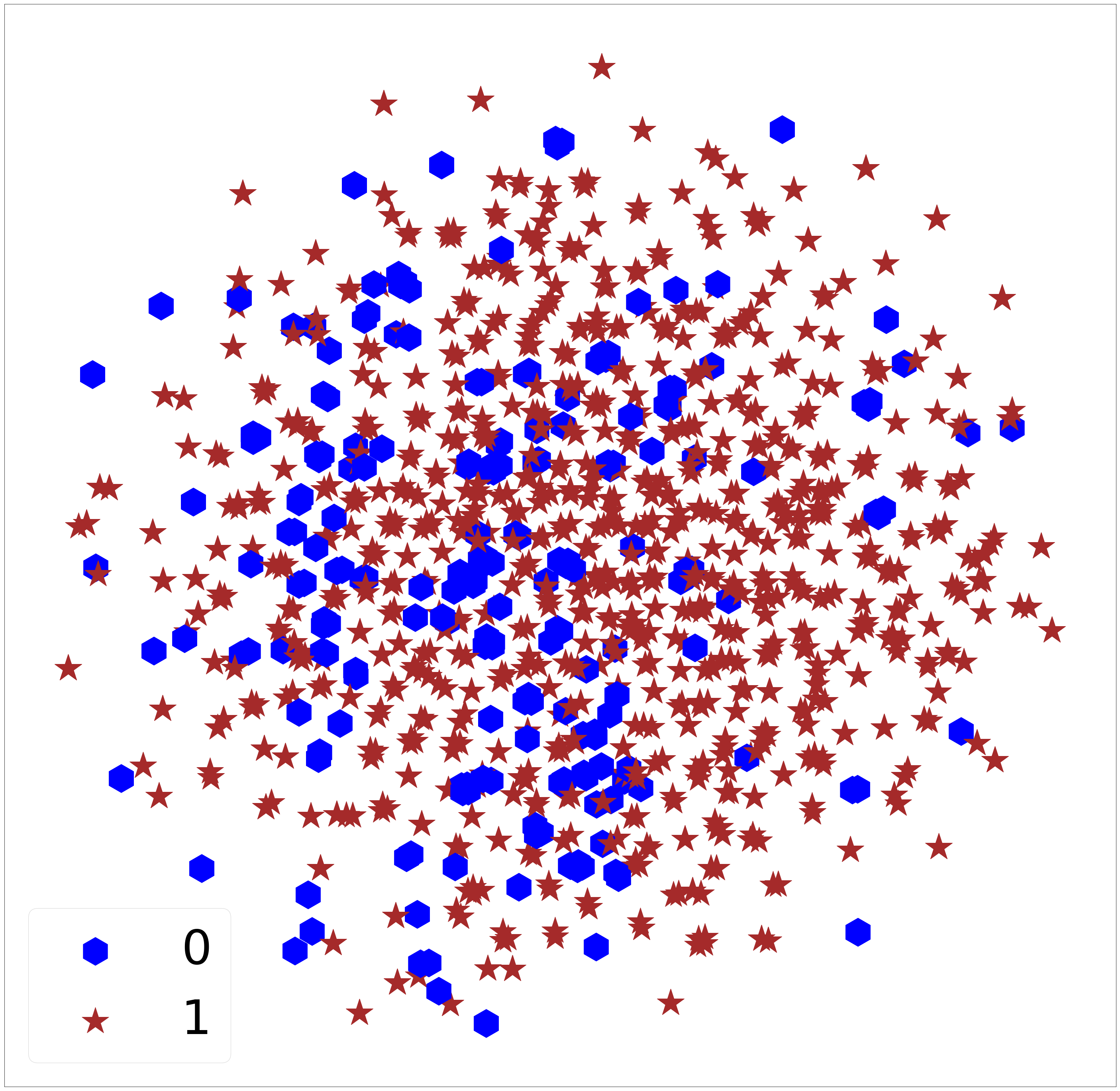}
\caption{$e_2$ embedding}
\end{subfigure}
\caption{\label{fig:tsne_german}German dataset -- t-SNE visualization. Labels indicate the biasing factor -- gender. Raw data clusters by gender, showing the bias. $e_1$ does not cluster by gender, as desired for fairness, while $e_2$ does, showing the migration of bias to $e_2$.}
\end{figure*}
}

\paragraph*{Adult} This is an income dataset of 45,222 individuals with various socio-economic attributes. The prediction task is to infer whether a person has more than \$50,000 savings. The biasing factor $z$ for this dataset is age, which is binarized, and it is required to make age-invariant savings predictions. We model the encoder and the $z$-discriminator as two-layer neural networks, and the predictor, the decoder, and the disentanglers as one-layer neural networks.

Results of this experiment are presented in Table~\ref{tab:adult}. Our model achieves the state-of-the-art performance at the accuracy of predicting $y$, while being completely invariant to $z$ as reflected by $A_z$ which is the same as the population share of the majority $z$-class ($0.67$). Figure~\ref{fig:tsne_adult} shows the t-SNE visualization of the raw data and the $e_1$ and $e_2$ embeddings. Both the raw data and the $e_2$ embedding show clustering by age while the invariant embedding $e_1$ does not.

\paragraph*{German} This dataset contains information about 1,000 people with the target to predict whether a person has a good credit-rating. The biasing factor here is gender and it is required to make gender-invariant credit assessments. For evaluating UnifAI on this dataset, the $z$-discriminator is modeled as a two-layer neural network whereas one-layer neural networks are used to instantiate the encoder, the predictor, the decoder, and the disentanglers.

\setlength{\tabcolsep}{1.75em} 
\begin{table*}
\makegapedcells
\centering
\caption{\label{tab:german}Results on German dataset}
\begin{tabular}{ c ?{1.5pt} c ?{0.75pt} c ?{0.75pt} c ?{0.75pt} c ?{1.5pt} c }
  \hbline
  \textbf{Metric} & \textbf{NN+MMD~\cite{bib:nnmmd}} & \textbf{VFAE~\cite{bib:vfae}} & \textbf{CAI~\cite{bib:sai}} & \textbf{CVIB~\cite{bib:cvib}} & \textbf{UnifAI (ours)} \\
  \hbline
  $A_y$ & 0.74 & 0.70 & 0.70 & 0.74 & \textbf{0.78} \\
  $A_z$ & \textbf{0.80} & \textbf{0.80} & 0.81 & \textbf{0.80} & \textbf{0.80} \\
  \hbline
\end{tabular}
\end{table*}
\setlength{\tabcolsep}{\defaulttabcolsep} 

Table~\ref{tab:german} summarizes the results of this experiment, showing that the proposed model outperforms previous methods at $A_y$, while retaining $A_z$ at the population share of the majority gender class (0.80). Thus, the proposed model achieves perfect invariance to gender while retaining more information relevant for making credit assessments. Figure~\ref{fig:tsne_german} shows the t-SNE visualization of the raw data and the $e_1$ and $e_2$ embeddings. While the raw data and the $e_2$ embedding are clustered by gender, the fair embedding $e_1$ is not.

\vspace{5pt}
\noindent As evident from results on both the datasets, our invariance induction framework works effectively at the task of fair representation learning, exhibiting state-of-the-art results.

\subsection{Competition between prediction \& reconstruction}
\label{subsec:eta}

\begin{figure}
\centering
\includegraphics[width=0.485\textwidth]{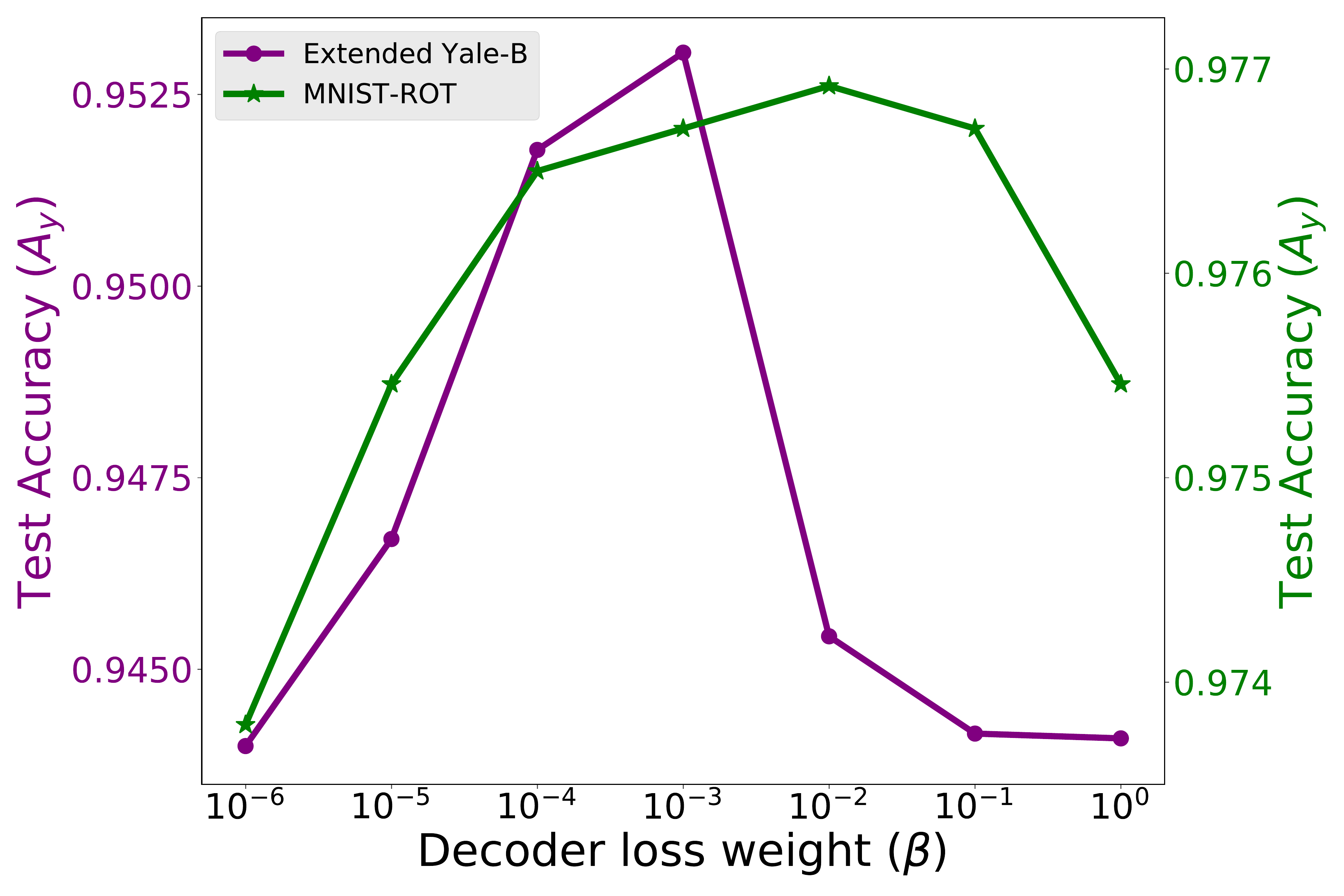}
\caption{\label{fig:eta_plot}Effect of competition between prediction and reconstruction on $y$-accuracy. Plots were generated by keeping $\alpha$ and $\gamma$ fixed at $100$ and $1$, respectively, and increasing $\beta$.}
\end{figure}

Figure~\ref{fig:eta_plot} shows the effect of the ratio $\frac{\alpha}{\beta}$ on the prediction performance ($A_y$) for the Extended Yale-B and MNIST-ROT datasets. The results were generated by keeping the loss-weights $\alpha$ and $\gamma$ fixed at $100$ and $1$, respectively, and increasing $\beta$ from $10^{-6}$ to $1$. Thus, the plots show the effect of gradually increasing the competition between the prediction and reconstruction tasks by giving the latter more say in the overall training objective. As evident, increasing $\beta$ improves $A_y$ by pulling nuisance factors into $e_2$ up to a point beyond which $A_y$ drops because information essential for predicting $y$ also gets pushed from $e_1$ to $e_2$. Hence, the observed behavior of the said competition is consistent with the intuitive analysis provided in Section~\ref{sec:theoretical}.


\section{Conclusion}
\label{sec:conclusion}

We have presented a unified framework for invariance induction in neural networks for both nuisance and biasing factors of data. Our method models invariance to nuisance as an information separation task, which is achieved by competitive training between a predictor and a decoder coupled with disentanglement, and explicitly penalizes the network if it encodes known biasing factors in order to achieve independence to such information. We described an adversarial instantiation of this framework and provided analysis of its working. Experimental evaluation shows that our invariance induction model outperforms state-of-the-art methods, which incorporate $z$-labels in their training, on learning invariance to nuisance factors without using any $z$-annotations. The proposed model also exhibits state-of-the-art performance on fairness tasks where it makes the latent embedding and the predictions independent of known biasing $z$. Our model does not make any assumptions about the data, and can, hence, be applied to any supervised learning task, eg., binary/multi-class classification or regression, without loss of generality.






%



\ifCLASSOPTIONcompsoc
  \section*{Acknowledgments}
\else
  \section*{Acknowledgment}
\fi

This work is based on research sponsored by the Defense Advanced Research Projects Agency under agreement number FA8750-16-2-0204. The U.S. Government is authorized to reproduce and distribute reprints for governmental purposes  notwithstanding any copyright notation thereon. The views and conclusions contained herein are those of the authors and should not be interpreted as necessarily representing the official policies or endorsements, either expressed or implied, of the Defense Advanced Research Projects Agency or the U.S. Government.

\ifCLASSOPTIONcaptionsoff
  \newpage
\fi



\bibliographystyle{IEEEtran}
\bibliography{IEEEabrv,main}

\begin{thebibliography}{10}
\providecommand{\url}[1]{#1}
\csname url@samestyle\endcsname
\providecommand{\newblock}{\relax}
\providecommand{\bibinfo}[2]{#2}
\providecommand{\BIBentrySTDinterwordspacing}{\spaceskip=0pt\relax}
\providecommand{\BIBentryALTinterwordstretchfactor}{4}
\providecommand{\BIBentryALTinterwordspacing}{\spaceskip=\fontdimen2\font plus
\BIBentryALTinterwordstretchfactor\fontdimen3\font minus
  \fontdimen4\font\relax}
\providecommand{\BIBforeignlanguage}[2]{{%
\expandafter\ifx\csname l@#1\endcsname\relax
\typeout{** WARNING: IEEEtran.bst: No hyphenation pattern has been}%
\typeout{** loaded for the language `#1'. Using the pattern for}%
\typeout{** the default language instead.}%
\else
\language=\csname l@#1\endcsname
\fi
#2}}
\providecommand{\BIBdecl}{\relax}
\BIBdecl

\bibitem{bib:feat_select_survey}
\BIBentryALTinterwordspacing
J.~Miao and L.~Niu, ``A survey on feature selection,'' \emph{Procedia Computer
  Science}, vol.~91, pp. 919 -- 926, 2016, promoting Business Analytics and
  Quantitative Management of Technology: 4th International Conference on
  Information Technology and Quantitative Management (ITQM 2016). [Online].
  Available:
  \url{http://www.sciencedirect.com/science/article/pii/S1877050916313047}
\BIBentrySTDinterwordspacing

\bibitem{bib:data_aug_1}
T.~Ko, V.~Peddinti, D.~Povey, and S.~Khudanpur, ``Audio augmentation for speech
  recognition,'' in \emph{Sixteenth Annual Conference of the International
  Speech Communication Association}, 2015.

\bibitem{bib:data_aug_2}
A.~Krizhevsky, I.~Sutskever, and G.~E. Hinton, ``Imagenet classification with
  deep convolutional neural networks,'' in \emph{Advances in neural information
  processing systems}, 2012, pp. 1097--1105.

\bibitem{bib:data_aug_3}
I.~Saito, J.~Suzuki, K.~Nishida, K.~Sadamitsu, S.~Kobashikawa, R.~Masumura,
  Y.~Matsumoto, and J.~Tomita, ``Improving neural text normalization with data
  augmentation at character- and morphological levels,'' in \emph{Proceedings
  of the Eighth International Joint Conference on Natural Language Processing
  (Volume 2: Short Papers)}.\hskip 1em plus 0.5em minus 0.4em\relax Asian
  Federation of Natural Language Processing, 2017, pp. 257--262.

\bibitem{bib:uai}
A.~Jaiswal, R.~Y. Wu, W.~Abd-Almageed, and P.~Natarajan, ``Unsupervised
  adversarial invariance,'' in \emph{Advances in Neural Information Processing
  Systems 31}, S.~Bengio, H.~Wallach, H.~Larochelle, K.~Grauman,
  N.~Cesa-Bianchi, and R.~Garnett, Eds.\hskip 1em plus 0.5em minus 0.4em\relax
  Curran Associates, Inc., 2018, pp. 5097--5107.

\bibitem{bib:infodropout}
A.~Achille and S.~Soatto, ``Information dropout: Learning optimal
  representations through noisy computation,'' \emph{IEEE Transactions on
  Pattern Analysis and Machine Intelligence}, vol.~40, no.~12, pp. 2897--2905,
  Dec 2018.

\bibitem{bib:vib}
A.~A. Alemi, I.~Fischer, J.~V. Dillon, and K.~Murphy, ``Deep variational
  information bottleneck,'' \emph{arXiv preprint arXiv:1612.00410}, 2016.

\bibitem{bib:drl}
Y.~Bengio, A.~Courville, and P.~Vincent, ``Representation learning: A review
  and new perspectives,'' \emph{IEEE transactions on pattern analysis and
  machine intelligence}, vol.~35, no.~8, pp. 1798--1828, 2013.

\bibitem{bib:face_rec}
I.~{Masi}, F.~{Chang}, J.~{Choi}, S.~{Harel}, J.~{Kim}, K.~{Kim}, J.~{Leksut},
  S.~{Rawls}, Y.~{Wu}, T.~{Hassner}, W.~{AbdAlmageed}, G.~{Medioni},
  L.~{Morency}, P.~{Natarajan}, and R.~{Nevatia}, ``Learning pose-aware models
  for pose-invariant face recognition in the wild,'' \emph{IEEE Transactions on
  Pattern Analysis and Machine Intelligence}, vol.~41, no.~2, pp. 379--393, Feb
  2019.

\bibitem{bib:bias_face_recognition}
M.~Merler, N.~Ratha, R.~S. Feris, and J.~R. Smith, ``Diversity in faces,''
  \emph{arXiv preprint arXiv:1901.10436}, 2019.

\bibitem{bib:bias_sentiment_analysis}
S.~Kiritchenko and S.~M. Mohammad, ``Examining gender and race bias in two
  hundred sentiment analysis systems,'' \emph{arXiv preprint arXiv:1805.04508},
  2018.

\bibitem{bib:bias_socioeconomic}
R.~{Courtland}, ``{Bias detectives: the researchers striving to make algorithms
  fair},'' \emph{Nature}, vol. 558, pp. 357--360, Jun. 2018.

\bibitem{bib:lfr}
R.~Zemel, Y.~Wu, K.~Swersky, T.~Pitassi, and C.~Dwork, ``Learning fair
  representations,'' in \emph{Proceedings of the 30th International Conference
  on Machine Learning}, ser. Proceedings of Machine Learning Research,
  S.~Dasgupta and D.~McAllester, Eds., vol.~28, no.~3.\hskip 1em plus 0.5em
  minus 0.4em\relax Atlanta, Georgia, USA: PMLR, 17--19 Jun 2013, pp. 325--333.

\bibitem{bib:nnmmd}
Y.~Li, K.~Swersky, and R.~Zemel, ``Learning unbiased features,'' \emph{arXiv
  preprint arXiv:1412.5244}, 2014.

\bibitem{bib:vfae}
C.~Louizos, K.~Swersky, Y.~Li, M.~Welling, and R.~Zeme, ``The variational fair
  autoencoder,'' in \emph{Proceedings of International Conference on Learning
  Representations}, 2016.

\bibitem{bib:sai}
Q.~Xie, Z.~Dai, Y.~Du, E.~Hovy, and G.~Neubig, ``Controllable invariance
  through adversarial feature learning,'' in \emph{Advances in Neural
  Information Processing Systems 30}, I.~Guyon, U.~V. Luxburg, S.~Bengio,
  H.~Wallach, R.~Fergus, S.~Vishwanathan, and R.~Garnett, Eds.\hskip 1em plus
  0.5em minus 0.4em\relax Curran Associates, Inc., 2017, pp. 585--596.

\bibitem{bib:cvib}
D.~Moyer, S.~Gao, R.~Brekelmans, A.~Galstyan, and G.~Ver~Steeg, ``Invariant
  representations without adversarial training,'' in \emph{Advances in Neural
  Information Processing Systems 31}, S.~Bengio, H.~Wallach, H.~Larochelle,
  K.~Grauman, N.~Cesa-Bianchi, and R.~Garnett, Eds.\hskip 1em plus 0.5em minus
  0.4em\relax Curran Associates, Inc., 2018, pp. 9102--9111.

\bibitem{bib:hcv}
R.~Lopez, J.~Regier, M.~I. Jordan, and N.~Yosef, ``Information constraints on
  auto-encoding variational bayes,'' in \emph{Advances in Neural Information
  Processing Systems 31}, S.~Bengio, H.~Wallach, H.~Larochelle, K.~Grauman,
  N.~Cesa-Bianchi, and R.~Garnett, Eds.\hskip 1em plus 0.5em minus 0.4em\relax
  Curran Associates, Inc., 2018, pp. 6117--6128.

\bibitem{bib:achille2018}
A.~Achille and S.~Soatto, ``Emergence of invariance and disentanglement in deep
  representations,'' \emph{Journal of Machine Learning Research}, vol.~19,
  no.~50, pp. 1--34, 2018.

\bibitem{bib:gan}
I.~Goodfellow, J.~Pouget-Abadie, M.~Mirza, B.~Xu, D.~Warde-Farley, S.~Ozair,
  A.~Courville, and Y.~Bengio, ``Generative {Adversarial} {Nets},'' in
  \emph{Advances in neural information processing systems}, 2014, pp.
  2672--2680.

\bibitem{bib:feat_select}
J.~Miao and L.~Niu, ``A survey on feature selection,'' \emph{Procedia Computer
  Science}, vol.~91, pp. 919 -- 926, 2016, promoting Business Analytics and
  Quantitative Management of Technology: 4th International Conference on
  Information Technology and Quantitative Management (ITQM 2016).

\bibitem{bib:sbmrinet}
A.~Jaiswal, D.~Guo, C.~S. Raghavendra, and P.~Thompson, ``Large-scale
  unsupervised deep representation learning for brain structure,'' \emph{arXiv
  preprint arXiv:1805.01049}, 2018.

\bibitem{bib:ib}
N.~Tishby, F.~C. Pereira, and W.~Bialek, ``The information bottleneck method,''
  in \emph{37th Annual Allerton Conference on Communication, Control and
  Computing}, 1999, pp. 368--377.

\bibitem{bib:dropout}
N.~Srivastava, G.~Hinton, A.~Krizhevsky, I.~Sutskever, and R.~Salakhutdinov,
  ``Dropout: A simple way to prevent neural networks from overfitting,''
  \emph{The Journal of Machine Learning Research}, vol.~15, no.~1, pp.
  1929--1958, 2014.

\bibitem{bib:mmd}
A.~Gretton, K.~M. Borgwardt, M.~Rasch, B.~Sch\"{o}lkopf, and A.~J. Smola, ``A
  kernel method for the two-sample-problem,'' in \emph{Advances in Neural
  Information Processing Systems 19}, B.~Sch\"{o}lkopf, J.~C. Platt, and
  T.~Hoffman, Eds.\hskip 1em plus 0.5em minus 0.4em\relax MIT Press, 2007, pp.
  513--520.

\bibitem{bib:vae}
D.~P. Kingma and M.~Welling, ``Auto-encoding {Variational} {Bayes},'' in
  \emph{International {Conference} on {Learning} {Representations}}, 2014.

\bibitem{bib:hsic}
A.~Gretton, O.~Bousquet, A.~Smola, and B.~Sch{\"o}lkopf, ``Measuring
  statistical dependence with hilbert-schmidt norms,'' in \emph{Algorithmic
  Learning Theory}, S.~Jain, H.~U. Simon, and E.~Tomita, Eds.\hskip 1em plus
  0.5em minus 0.4em\relax Berlin, Heidelberg: Springer Berlin Heidelberg, 2005,
  pp. 63--77.

\bibitem{bib:mtl}
S.~Ruder, ``An overview of multi-task learning in deep neural networks,''
  \emph{arXiv preprint arXiv:1706.05098}, 2017.

\bibitem{bib:eyb}
A.~S. Georghiades, P.~N. Belhumeur, and D.~J. Kriegman, ``From few to many:
  illumination cone models for face recognition under variable lighting and
  pose,'' \emph{IEEE Transactions on Pattern Analysis and Machine
  Intelligence}, vol.~23, no.~6, pp. 643--660, Jun 2001.

\bibitem{bib:chairs}
M.~Aubry, D.~Maturana, A.~Efros, B.~C.~Russell, and J.~Sivic, ``Seeing 3d
  chairs: Exemplar part-based 2d-3d alignment using a large dataset of cad
  models,'' in \emph{Proceedings of the IEEE Computer Society Conference on
  Computer Vision and Pattern Recognition}, 06 2014.

\bibitem{bib:tsne}
L.~v.~d. Maaten and G.~Hinton, ``Visualizing data using t-sne,'' \emph{Journal
  of machine learning research}, vol.~9, no. Nov, pp. 2579--2605, 2008.

\bibitem{bib:mnist}
Y.~LeCun, L.~Bottou, Y.~Bengio, and P.~Haffner, ``Gradient-based {Learning}
  {Applied} to {Document} {Recognition},'' \emph{Proceedings of the IEEE},
  vol.~86, no.~11, pp. 2278--2324, 1998.

\bibitem{bib:gan_aug}
A.~Antoniou, A.~Storkey, and H.~Edwards, ``Data augmentation generative
  adversarial networks,'' \emph{arXiv preprint arXiv:1711.04340}, 2017.

\bibitem{bib:fashion_mnist}
H.~Xiao, K.~Rasul, and R.~Vollgraf. (2017) Fashion-mnist: a novel image dataset
  for benchmarking machine learning algorithms.

\bibitem{bib:omniglot}
B.~M. Lake, R.~Salakhutdinov, and J.~B. Tenenbaum, ``Human-level concept
  learning through probabilistic program induction,'' \emph{Science}, vol. 350,
  no. 6266, pp. 1332--1338, 2015.

\bibitem{bib:bicogan}
A.~Jaiswal, W.~AbdAlmageed, Y.~Wu, and P.~Natarajan, ``Bidirectional
  conditional generative adversarial networks,'' in \emph{Computer Vision --
  ACCV 2018}.\hskip 1em plus 0.5em minus 0.4em\relax Springer International
  Publishing, 2019.

\bibitem{bib:dann}
Y.~Ganin, E.~Ustinova, H.~Ajakan, P.~Germain, H.~Larochelle, F.~Laviolette,
  M.~Marchand, and V.~Lempitsky, ``Domain-adversarial training of neural
  networks,'' \emph{The Journal of Machine Learning Research}, vol.~17, no.~1,
  pp. 2096--2030, 2016.

\bibitem{bib:amzrev}
M.~Chen, Z.~Xu, K.~Q. Weinberger, and F.~Sha, ``Marginalized denoising
  autoencoders for domain adaptation,'' in \emph{Proceedings of the 29th
  International Conference on Machine Learning}, ser. ICML'12.\hskip 1em plus
  0.5em minus 0.4em\relax USA: Omnipress, 2012, pp. 1627--1634.

\bibitem{bib:adult}
R.~Kohavi, ``Scaling up the accuracy of naive-bayes classifiers: A
  decision-tree hybrid,'' in \emph{Proceedings of the Second International
  Conference on Knowledge Discovery and Data Mining}, ser. KDD'96.\hskip 1em
  plus 0.5em minus 0.4em\relax AAAI Press, 1996, pp. 202--207.

\bibitem{bib:uci}
\BIBentryALTinterwordspacing
D.~Dheeru and E.~Karra~Taniskidou, ``{UCI} machine learning repository,'' 2017.
  [Online]. Available: \url{http://archive.ics.uci.edu/ml}
\BIBentrySTDinterwordspacing

\end{thebibliography}

\end{document}